\documentclass[letterpaper, 10 pt, conference]{ieeeconf}  

\usepackage[table]{xcolor}

\usepackage{multirow}
\usepackage{times}

\usepackage{hyperref}

\usepackage{color}
\usepackage{algorithm}
\usepackage{algpseudocode}
\usepackage{listings}
\usepackage{courier}
\usepackage{soul}
\usepackage{amsmath,amsfonts,amssymb}

\usepackage{mathtools}
\usepackage{subfigure}
\usepackage{wrapfig}
\usepackage{hyperref}
\usepackage{multirow}
\usepackage{graphicx}
\usepackage{graphicx}
\usepackage{soul}
\usepackage[flushleft]{threeparttable}
\usepackage{enumerate}
\usepackage{algorithmicx}
\usepackage{algpseudocode}
\usepackage{epstopdf}

\IEEEoverridecommandlockouts                              
\definecolor{red}{rgb}{1.00,0.00,0.00}
\definecolor{blue}{rgb}{0.00,0.00,1.00}
\definecolor{green}{rgb}{0.2,0.70,0.2}
\definecolor{yellow}{rgb}{0.5,0.5,0.0}
\definecolor{white}{rgb}{1,1,1}

\title{\LARGE \bf
A Robust Closed-Loop Biped Locomotion Planner Based on \\Time Varying Model Predictive Control
	\author{Mohammadreza Kasaei, Nuno Lau and Artur Pereira
	 \\IEETA / DETI University of Aveiro 3810-193 Aveiro, Portugal \\
		\{mohammadreza, nunolau, artur\}@ua.pt
	}
}

\begin{document}
\maketitle
\thispagestyle{empty}
\pagestyle{empty}
\begin{abstract}
Developing robust locomotion for humanoid robots is a complex task due to the unstable nature of these robots and also to the unpredictability of the terrain. A robust locomotion planner is one of the fundamental components for generating stable biped locomotion. This paper presents an optimal closed-loop biped locomotion planner which can plan reference trajectories even in challenging conditions. The proposed planner is designed based on a Time-Varying Model Predictive Control~(TVMPC) scheme to be able to consider some constraints in the states, inputs and outputs of the system and also mixed input-output. Moreover, the proposed planner takes into account the vertical motion of the Center of Mass~(COM) to generate walking with mostly stretched knees which is more human-like. Additionally, the planner uses the concept of Divergent Component of Motion~(DCM) to modify the reference ZMP online to improve the withstanding level of the robot in the presence of severe disturbances. The performance and also the robustness of the proposed planner are validated by performing several simulations using~\mbox{MATLAB}. The simulation results show that the proposed planner is capable of generating the biped locomotion robustly.
\end{abstract}

\section{Introduction}
\label{sec:introduction}
Despite years of research in developing robust locomotion for humanoid robots, this subject is still an active topics in the robotics community. Humans expect to adapt this type of robot in society to be really useful in performing daily-life tasks. One of the fundamental requirements to achieve this is performing stable walking in challenging situations, like walking on uneven terrains or regaining balance after suffering an unpredictable disturbance.
A common approach in the literature for developing walking is to generate the COM trajectory using preview control of the Zero Moment Point~(ZMP)~\cite{vukobratovic1970stability} according to a set of predefined footsteps and a simplified dynamics model~(e.g. Linear Inverted Pendulum Model~(LIPM)~\cite{kajita2003biped,shimmyo2013biped,kasaei2018optimal}). In these approaches, the footstep planner generates a set of footsteps without taking into account the current state of the COM. Wieber~\cite{wieber2006trajectory} proposed a new scheme based on Model Predictive Control~(MPC) and showed how this scheme improves the original ZMP Preview Control scheme. Pratt et al.~\cite{pratt2006capture} proposed the Capture Point~(CP) concept which uses the current state of the COM to determine a point on the ground that the robot should step to keep its stability. Later, Takaneka et al.~\cite{takenaka2009real} proposed the Divergent Component of Motion~(DCM) concept and used it to real-time plan and control of a humanoid walking. Englsberger et al.~\cite{englsberger2015three} extended DCM to 3D and introduced Enhanced Centroidal Moment Pivot point~(eCMP) and also the Virtual Repellent Point~(VRP) which could be used to encode the direction, magnitude and total forces of the external push. Khadiv et al.~\cite{khadiv2016stepping} proposed a method based on combining DCM tracking and step adjustment to stabilize biped locomotion. Griffin~\cite{griffin2017walking} et al. developed a walking stabilization system using step time and location adjustment based on DCM concept and showed the performance of their method using simulations and experiments. Hopkins et al.~\cite{hopkins2014humanoid} introduced time-varying DCM and showed that by varying the natural frequency of the DCM generic COM height trajectories during stepping can be achieved. Kryczka et al.~\cite{kryczka2015online} developed an algorithm for online locomotion planning based on a nonlinear optimization technique. Their method finds a set of step parameters which controls the robot from the current state to the target state. Herdt et al.~\cite{herdt2010online} showed that walking based on an automatic footstep placement can be formulated as an MPC problem. Later, Herdt et al.~\cite{herdt2010walking} extended this work to enhance the flexibility and independency by defining polygonal constraints on the positions of the computed feet positions. Mitsuharu et al.~\cite{morisawa2005pattern} developed a biped locomotion planner which takes into account the vertical motion of the COM by defining a parametric surface to determine the relative height of the COM from the landing foot position. The effectiveness of their method has been verified by stair climbing simulation and walking with mostly stretched knee in real experiment. Recently, Kajita et al.~\cite{kajita2018biped} proposed a method based on Spatially Quantized Dynamics~(SQD) to control biped locomotion. They showed by moving the COM up and down a few centimeters at each step, robot could walk with stretched support legs. The validity of their method has been confirmed by performing real experiments using HRP-2Kai.
In most of the aforementioned works, the height of COM is considered to be fixed and its motion is restricted to move along a horizontal defined plane. 

In this paper, we release this constraint and formulate the problem of developing an online optimal closed-loop biped locomotion planner as a time-varying MPC. This scheme tries to predict how the dynamics model varies over the prediction horizon and we will show that by previewing the dynamics model changes the performance of the planner will be improved. Besides, the presented scheme uses the concept of DCM to plan the ZMP reference trajectory to increase the withstanding level of the robot in the presence of severe disturbances. We have performed some simulations to show the robustness of the planner in challenging situations. The remainder of this paper is organized as follows: Section~\ref{sec:DynamicsModel} gives an overview of the dynamics model and explain how the output of the system can be predicted. The fundamentals of the proposed MPC planner is explained in Section~\ref{sec:MPCPlanner}. Afterward in Section~\ref{sec:TVMPC}, it will be extended to a TVMPC in order to release the height constraint of the LIPM. The simulation scenarios and the results will be presented in Section~\ref{sec:simulation} and, based on the simulation results, discussion and comparison will be given in Section~\ref{sec:discussion}. Finally, conclusions and future research are presented in Section~\ref{sec:CONCLUSION}.
\vspace{-2mm}
\section{Dynamics Model}
\label{sec:DynamicsModel}
In this section, a brief review of LIPM and its properties will be presented and then we will discretize this dynamics model and use it as the fundamental of our future prediction approach. Furthermore, we will explain how the output of the system can be predicted for a given set of control inputs. It should be noted that, since all the equations in sagittal and frontal planes are equivalent and independent, we derive the equations just in the sagittal plane.

\subsection{Linear Inverted Pendulum Model~(LIPM)}
LIPM abstracts the dynamics model of a humanoid robot by concentrating the whole robot mass in its COM which is connected to the ground via a mass-less rod. The length of the rod is equal to the height of COM. To decouple the motion in sagittal and frontal planes, the mass is limited to move along a predefined horizontal plane. According to these assumptions, the overall dynamics of a humanoid robot can be represented by a first-order stable dynamics as follow:
\begin{equation}
\ddot{x_c} = \omega^2 ( x_c - p) \quad ,
\label{eq:lipm}
\end{equation}
\noindent
where $x_c$ represents the position of COM, \mbox{$\omega = \sqrt{\frac{g+\ddot{z}_c}{z_c-p}}$} is the natural frequency of the pendulum, $g$ represents the gravity acceleration, $z_c$ is the height of COM and $p$ is the position of ZMP which is a point on the ground plane where the ground reaction force acts to compensate gravity and inertia~\cite{vukobratovic1970stability}. 

Using (\ref{eq:lipm}) and by considering the COM jerk as a manipulated variable, the dynamics model can be represented as a state space system as follows:
\begin{equation}
\frac{d}{dt} \begin{bmatrix} x_c \\ \dot{x}_c \\ \ddot{x}_c \end{bmatrix}
= 
\begin{bmatrix} 
0 & 1 & 0  \\
0 & 0 & 1 \\
0 & 0 & 0 
\end{bmatrix}	
\begin{bmatrix} x_c \\ \dot{x}_c \\ \ddot{x}_c \end{bmatrix}
+
\begin{bmatrix} 
0 \\
0 \\
1 
\end{bmatrix}  \dddot{x}_c \quad,
\label{eq:statespace_lipm}
\end{equation}
where $\dddot{x}_c$ represents the jerk of COM in $X$ direction. This state space system represents a linear continuous system which should be discretized in order to discrete time implementation. To do that, the jerk command is considered to be kept within a time step~(T). Moreover, based on (\ref{eq:lipm}), ZMP can be measured at each control cycle. Thus, the discrete system is obtained as follows:
\begin{equation}
\begin{gathered} 
X(k+1)=AX(k)+Bu(k)\quad \\
y(k) = CX(k)
\end{gathered} 
\label{eq:statespace_lipm_disc} 
\end{equation}
\noindent
where
\begin{equation}
\begin{aligned} 
X &= \begin{bmatrix} x_c & \dot{x}_c & \ddot{x}_c \end{bmatrix}^\top &&,\\
A &= \begin{bmatrix} 
1 & T & T^2/2 \\
0 & 1 & T \\
0 & 0 & 1
\end{bmatrix}\quad &&, \quad
B = \begin{bmatrix} T^3/6 \\ T^2/2 \\ T  \end{bmatrix}, \\
C &= \begin{bmatrix} 1 & 0 & -\omega^{2}  \end{bmatrix} &&, \quad 
u = \dddot{x}_c  \quad . 
\end{aligned} 
\label{eq:lipm_disc2}
\end{equation}

\subsection{Output Prediction}
According to (\ref{eq:statespace_lipm_disc}) and assuming that at the current sampling instant $k$ the state vector $X(k)$ is available through estimation based on a Kalman Filter~(KF), for $N_c$ (control horizon) assumed future control moves ($\Delta U =[ \Delta u(k), \Delta u(k+1),...,  \Delta u(k+N_c-1)]^\top$), the future behavior of the system ($Y = [y(k+1|k),y(k+2|k),...,y(k+N_p|k)]^\top$) can be predicted over a prediction horizon of $N_p$ time steps by applying this equation sequentially. It can be shown that all the predicted variables are formulated in terms of current state $X(k)$ and the future control movements $\Delta U$ as follows:
\begin{equation}
Y = FX(K_i)+\Phi \Delta U
\label{eq:ComplexForm_Predictive}
\end{equation}
\noindent 
where 
\begin{equation}
\resizebox{0.98\linewidth}{!}{$
\begin{gathered} 
F = \begin{bmatrix} 
CA &
CA^2&
CA^3&
\cdots&
CA^{N_p} 
\end{bmatrix} ^\top , \\
\Phi = 
\begin{bmatrix} 
 CB &0& 0& \cdots& 0\\
 CAB &CB& 0& \cdots& 0\\
 CA^2B &CAB& CB& \cdots& 0\\
 \vdots&\vdots&\vdots&\vdots&\vdots\\
 CA^{N_{p}-1}B & CA^{N_{p}-2}B& CA^{N_{p}-3}B&\cdots& CA^{N_{p}-N_c}B
 \end{bmatrix} .
\end{gathered} $ }
\label{eq:ComplexForm_matrix}
\end{equation} 
It should be noted that in case of $N_c < N_p$, the last controller output is kept for the remaining prediction horizon. In the next section, we will explain how this discrete system and the prediction procedure will be used to design and implement an MPC for planning the walking reference trajectories.

\section{Online Reference Planning Using MPC}
\label{sec:MPCPlanner}
As it is explained, for a given set of control inputs, the outputs of the system can be predicted within a prediction horizon. Based on this prediction, an MPC is used to determine the set of control inputs to keep the system outputs at or near specified reference values. Indeed, MPC is a practical framework which uses the current state of the system as an initial condition and solves the problem of finding an optimal set of control inputs~($\Delta U$) that minimize a specific cost function using quadratic programming~(QP). After determining the set of control inputs, only the first control signal~($\Delta u(0)$) is applied to the system and at the next control cycle, based on the new output measurements, MPC re-calculate a new set of control inputs. Although the set of control inputs is determined in an open-loop manner but re-calculating them at each control cycle turns this open-loop controller into an implicit closed-loop controller.

\subsection{Cost Function}
In our target framework, MPC deals with the problem of finding an optimal set of control inputs for a given system such that the outputs of the system keeps tracking the references. Thus, the cost function can be defined as follows:
\begin{equation}
 J(z_k) = \sum_{i=1}^{N_p}\{(r(k+i|k) - y(k+i|k))\}^2 
\label{eq:costfunction}
\end{equation}
\noindent
where $k$ represents current control interval, \mbox{$z_k^\top = \{\Delta u(k|k)^\top\quad \Delta u(k+1|k)^\top  ... \Delta u(k+N_p-1|k)^\top \}$} is QP decision, $r(k+i|k)$ is the reference value for the system outputs at $i^{th}$ prediction horizon step.
\subsection{Constraints}
Although the proposed online reference planning is already able to plan the locomotion trajectories, it is not completed yet because the generated plan might be not applicable. One of the important advantages of MPC-based planner is its ability to consider some constraints. To generate a feasible plan, MPC considers a set of constraints such that any plan which is not satisfying these constraints cannot be generated. Generally, a set of mixed input/output constraints can be specified in the following form:
\begin{equation}
Eu(k + j|k) + Fy(k + j|k) \le G + \epsilon
\label{eq:constraints}
\end{equation}
\noindent
where $j = 0, 1,...,N_p$, $k$ represents current time, $E, F, G$ are time variant matrices where each row of them represents a linear constraint. $\epsilon$ is used to define a slack variable to soft constraints. Moreover, based on this equation, we can also bound the inputs and the outputs to specified limitations. The most important constraint that our planner should consider is keeping the ZMP (output of the system) within the support polygon to ensure walking stability. This constraint can be established using the following parameters: 
\begin{equation}
\begin{aligned}
E &=[0\quad 0]^\top, \quad F=[1\quad-1]^\top, \\ G &= [\frac{0.9L_f}{2}\quad-\frac{0.9L_f}{2}]^\top, \quad \epsilon = [0\quad 0]^\top ,
\end{aligned}
\label{eq:param_constraint}
\end{equation}
\noindent
where $L_f$ represents the length of the robot's foot. It should be noted that we consider $90\%$ of the foot size to prevent ZMP from being too close to the edges of support polygon. Moreover, this constraint is considered to be a hard constraint which means the solution must always satisfy it.

\section{Release the Height Constraint of LIPM Using Time-Varying MPC}
\label{sec:TVMPC}
As it is explained before, LIPM considers an assumption to restrict the vertical motion of COM along a predefined plane. According to this assumption, a robot should always keep bending its knees which needs high torques in the ankle, knee and hip joints. Hence, generating a walking with this style does not only look unnatural but also it consumes more energy. To overcome this limitation and generate more energy efficient and also more human-like motion, we release this constraint by formulating the problem as a Time-Varying MPC~(TVMPC). Actually, this method is based on the prediction of how the dynamics model varies over the prediction horizon. 

\subsection{Time-Varying Prediction}
%
By releasing the height constraint of COM, (\ref{eq:statespace_lipm_disc}) will be changed to a time-varying model and the future behavior of the system can be predicted over a prediction horizon as follows:
\begin{equation}
\begin{bmatrix} y(k+1) \\\vdots\\ y(k+N_p) \end{bmatrix} =
S_xx(0)+S_{u1}u(-1)+S_u\begin{bmatrix} \Delta u(0) \\ \vdots \\ \Delta u(N_p-1) \end{bmatrix} 
\label{eq:pridection_tv},
\end{equation}
\noindent
where 
\begin{equation}
\resizebox{0.95\linewidth}{!} {$
\begin{aligned}
S_x &= \begin{bmatrix} C(1)A(0) \\C(2)A(1)A(0)\\ \vdots\\ C(N_p) \prod_{i=0}^{N_p-1}A(i) \end{bmatrix}, \\
S_{u1} &= \begin{bmatrix} C(1)B(0) \\C(2)[B(1)+A(1)B(0)]\\ \vdots \\ C(N_p) \sum_{k=0}^{N_p-1}[(\prod_{i=k+1}^{N_p-1}A(i))B(k)] \end{bmatrix}, \\
S_u &= \begin{bmatrix} 0 & 0& \hdots & 0\\
S_{u1} C(2)B(1)& 0& \hdots & 0\\
\vdots\\
 C(N_p)\sum_{k=1}^{N_p-1}[(\prod_{i=k+1}^{N_p-1}A(i))B(k)]&\hdots & \hdots &C(N_p)B(N_p-1) \end{bmatrix}. 
\end{aligned}$}
\label{eq:param_pridection_tv}
\end{equation}
According to this prediction, the problem of finding an optimal set of control inputs to track the reference trajectories can be formulated as a QP by considering a cost function and a set of constraints as were described in the previous section.

\subsection{Adaptive Reference Planning and Previewing the Reference and the Dynamics Model}
In our target framework, the procedure of reference planning is started by specifying the next foot step and based on that the reference ZMP will be generated. Indeed, a normal step is composed of two phases, which are single support~(SS) and double support~(DS). In order to achieve a stable walking, ZMP should be kept in the middle of support foot during SS phase and it should be moved proportionally to the COM during DS phase. Thus, a static ZMP reference generator can be formulated as follows:
\begin{equation}
\begin{aligned}
p (t)= 
\begin{cases}
f_{i}  &0 \leq t < T_{ss} \\
f_{i}+  \frac{(t-T_{ss})}{T_{ds}}S_L &T_{ss} \leq t < T_{ss}+T_{ds} 
\end{cases} 
\end{aligned},
\label{eq:zmpEquation}
\end{equation}
\noindent
where $t$ is current time which will be reset at the end of each step ($t \geq T_{ss}+T_{ds}$), $T_{ss}$ and $T_{ds}$ represent the duration of SS and DS phases, respectively, $S_L$ denotes the step length, $f_{i}$ represents a set of planned foot positions on a 2D surface ($i \in \mathbb{N}$). This formulation does not consider the current state of the COM and plans the ZMP at the beginning of each step. Therefore, to achieve an adaptive ZMP planner which considers the current state of the system, the concept of DCM is used to adjust the landing location of the swing leg~(next step location). Conceptually, DCM is the unstable part of COM dynamics which is a point that robot should step to come to rest~\cite{pratt2006capture} and it can be defined using a first-order stable dynamics as follows:
\begin{equation}
\zeta = x + \frac{\dot{x}}{\omega} \quad ,
\label{eq:dcm}
\end{equation}
\noindent
where $\zeta$ represents the DCM. According to the observability of state of the system at each control cycle and by solving (\ref{eq:dcm}) as an initial value problem, the landing location of the swing leg can be predicted in advance~\cite{takenaka2009real,englsberger2015three,englsberger2017smooth,kamioka2018simultaneous}. Therefore, to have an adaptive ZMP planner, step length can be adjusted at each control cycle as follows:
\begin{equation}
S_L (t) =  (\zeta_t - f_{i})e^{w(T_{ss}+T_{ds}-t)} \quad .
\label{eq:dcm_at_end}
\end{equation}
\begin{figure}[!t]
	\centering
	\includegraphics[width = 0.95\columnwidth, trim= 0.0cm 3cm 0cm 1.8cm,clip]{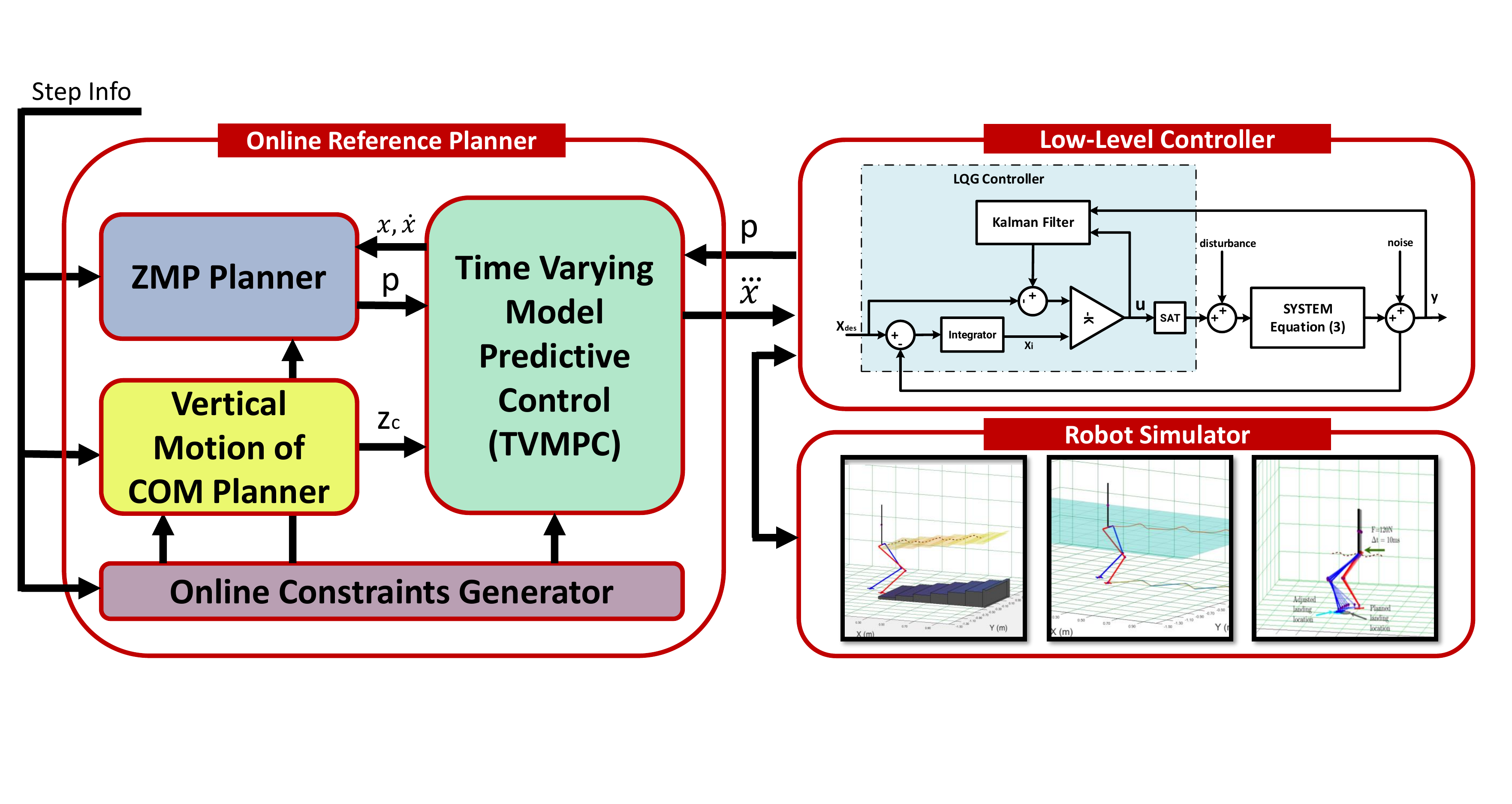}
	\vspace{-0mm}
	\caption{ The overall architecture of the proposed online planner which is composed of four main modules: ZMP planner, vertical motion of COM planner, online constraints generator and TVMPC.}
	\vspace{-5mm}
	\label{fig:overall}
\end{figure}
It should be mentioned that a compliance margin is defined to prevent unnecessary adjustment. Moreover, the new landing position is saturated inside a kinematically reachable area of the robot. According to this adaptive ZMP planner, we know how the setpoint will change in future and based on~(\ref{eq:costfunction}), by previewing the setpoint during the prediction horizon, the performance of the planner will be improved. Moreover, the performance of the planner will be improved more by previewing the changes of dynamics model during the prediction horizon. According to~(\ref{eq:pridection_tv}), if the vertical trajectory of the height of COM can be planned in advance, then the changes of dynamics model is predictable~\cite{morisawa2005pattern},~\cite{kajita2018biped}. Thus, we consider that a specific function can be defined to plan the vertical trajectory of COM during SS and also DS phase. During SS phase, the vertical motion of the COM is generated by a sinusoidal function and during the DS phase, the COM moves proportionally to its new position at the next step. The trajectory of the vertical motion of COM can be generated using the following function:
\begin{equation}
\resizebox{0.95\linewidth}{!} {$
\begin{aligned}
z_c(t)= 
\begin{cases}
z_{c0}+A_{ss}\sin(\frac{\pi}{T_{ss}}t)  &0 \leq t < T_{ss} \\
z_{c0}+ A_{ds}\sin(\frac{\pi}{T_{ds}}(t-T_{ss}))+ \frac{\Delta z_c}{T_{ds}}(t-T_{ss}) &T_{ss} \leq t < T_{ss} +T_{ds} 
\end{cases} 
\end{aligned},$}
\label{eq:z_traj}
\end{equation}
where $z_{c0}$ denotes the initial height of COM, $A_{ss}$ and $A_{ds}$ are the amplitudes of sinusoidal functions in SS and DS, respectively, $\Delta z_c$ represents the difference between the height of COM at the beginning and the end of each step. Using this vertical motion trajectory, a more human-like and more efficient walking with mostly stretched knees can be generated.

\begin{figure}[!t]
	\centering
	\begin{tabular}	{c c}			
		\includegraphics[width=0.45\linewidth, trim= 0cm 0cm 0cm 0cm,clip]{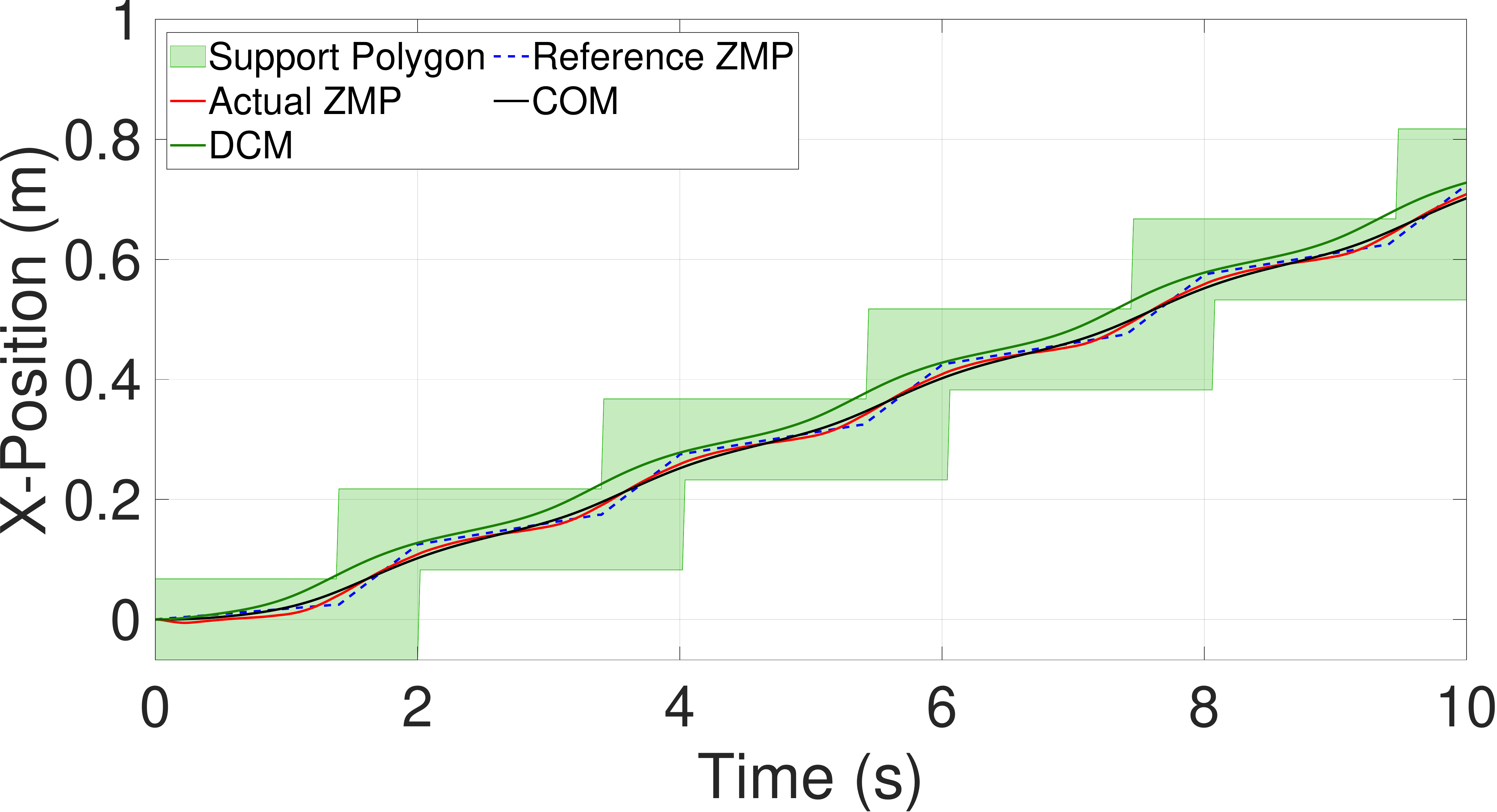}&
		\includegraphics[width=0.45\linewidth, trim= 0cm 0cm 0cm 0cm,clip]{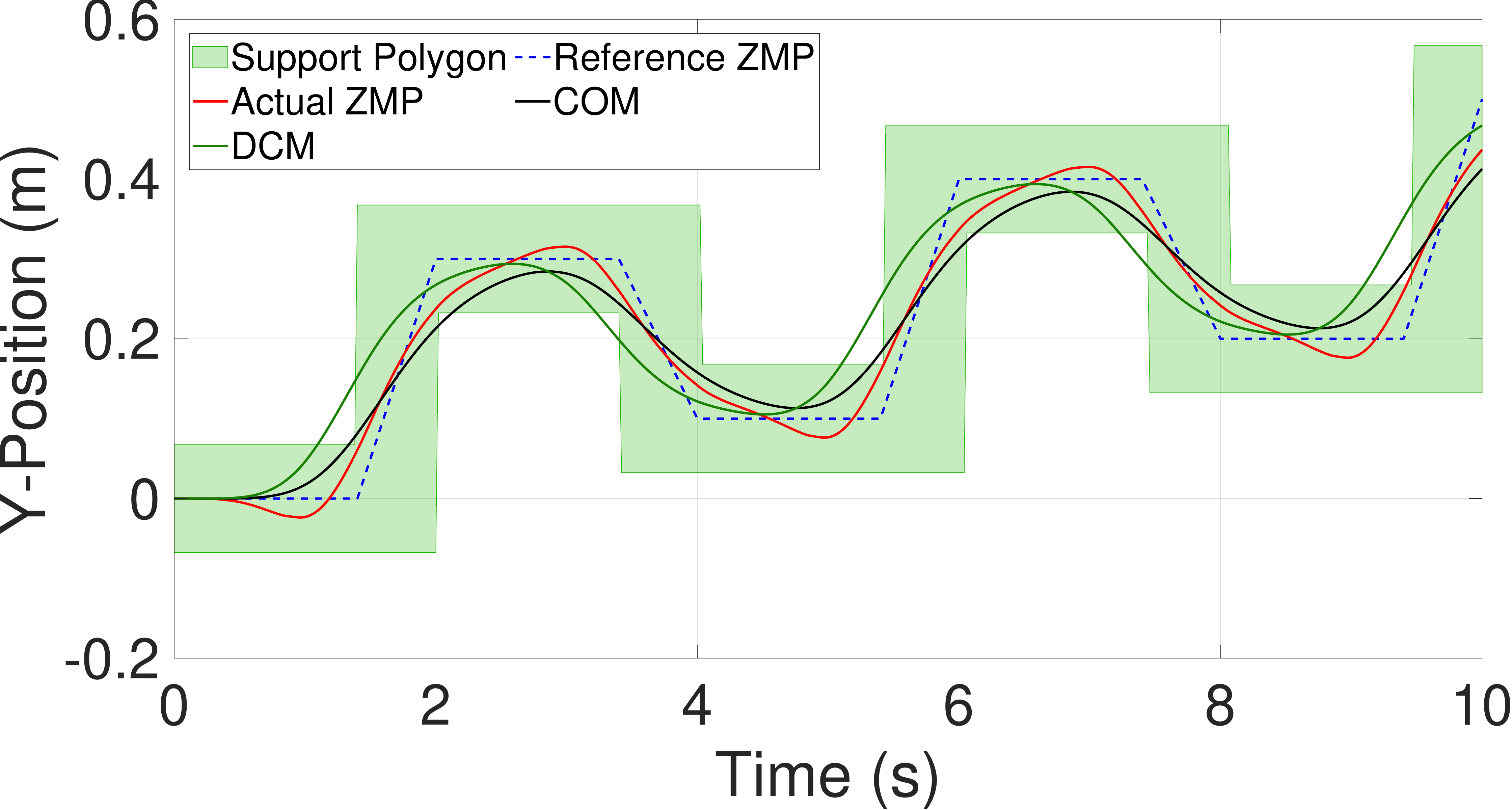}\\
		\hspace{-2mm}\includegraphics[width=0.45\linewidth, trim= 0cm 0cm 0cm 0cm,clip]{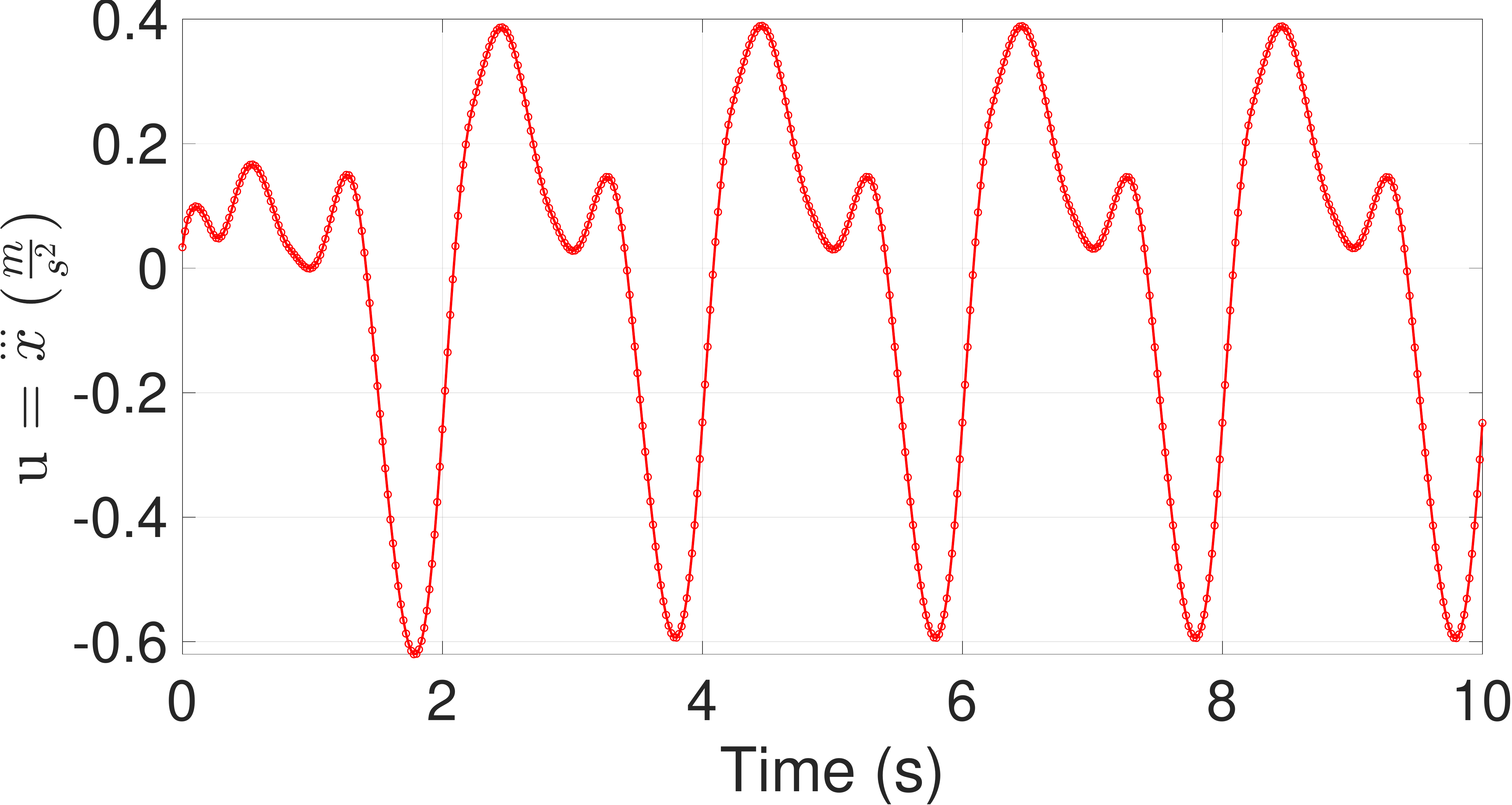}&
		\includegraphics[width=0.45\linewidth, trim= 0cm 0cm 0cm 0cm,clip]{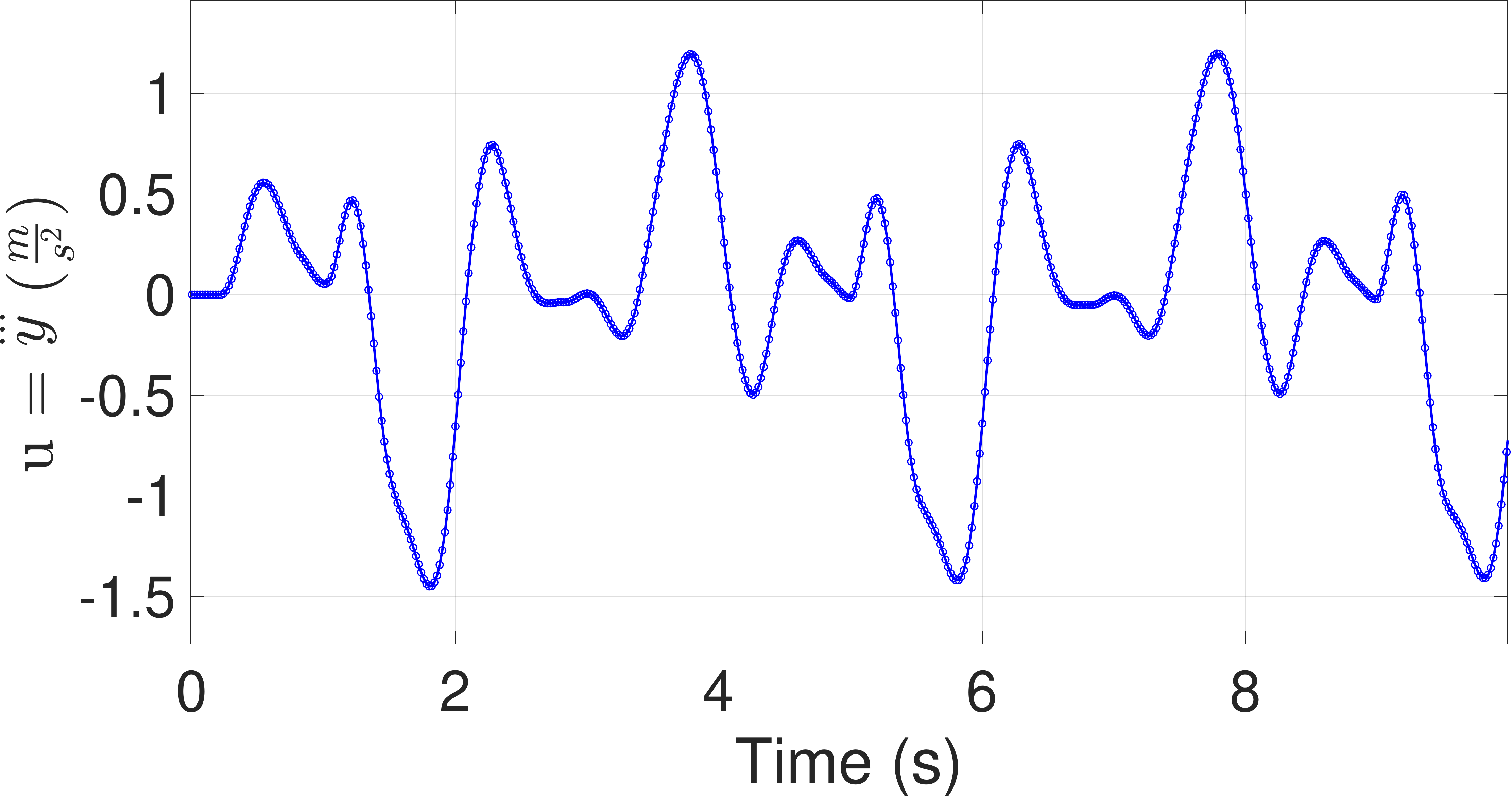}
	\end{tabular}	
	\vspace{-2mm}
	\caption{ The simulation results of diagonal walking scenario. The plots in the first row represent the reference ZMP, the measured ZMP and the measured COM position in X and Y directions. The plots in the second row show how the planner manipulates the jerk commands to keep the ZMP inside the support polygon and near the reference ZMP.}
	\vspace{-5mm}			
	\label{fig:diag_walking}
\end{figure}

\section{Simulation Results}
\label{sec:simulation}
This section will be started by an overview of the overall architecture of the proposed reference planner. Afterward, a set of simulations using a simulated robot will be performed to verify the performance of the planner. In these simulations, a humanoid robot is simulated using \mbox{MATLAB} based on the dynamics model explained in Section~\ref{sec:DynamicsModel}. The weight of the simulated robot is considered to be $100kg$, the height of COM is assumed to be $1m$ and both the foot width and the foot length are $0.075m$. Finally, the robustness of the planner will be analyzed with regarding to measurement noise, COM height estimation error and external disturbance. 

\subsection{Overall Architecture of the Planner}
The overall architecture of the proposed planner is depicted in Fig.~\ref{fig:overall}. As shown in this figure, it is composed of four main modules which are the ZMP planner, the vertical motion planner of COM, the online constraints generator and the TVMPC. The TVMPC planner uses the ZMP and the vertical trajectory of the COM as its reference inputs and, using the measured ZMP signal and the input constraints, plans the COM and the jerk command for the low-level controller which is responsible for controlling full body motion based on the generated command. Indeed, the detail of the low-level controller is beyond the scope of this paper and we assumed that it is able to track the reference trajectories robustly. 
\begin{table}[h!]
	\centering
	\caption{Parameters used in the simulations.}
	\label{tb:Params}	
	\resizebox{\linewidth}{!}{
		\begin{tabular}{c | c |c| c | c |c | c| c | c|c|c|c} 		
			$S_w$&$S_L$ &  $T_{ss}$ & $T_{ds}$ & $L_f$&$z_{c0}$ & $A_{ss}$ & $A_{ds}$ & $\Delta z_c$ & $T_s$&$N_p$&$N_c$ \\ 
			\hline			
			$0.1m$ & $0.2m$ & $1.5s$ & $0.5s$ & $0.075m$ & $1m$ & $0.0135m$ & $0.00135m$ & $0$ & $0.02s$ & $50$ & $2$ 
		\end{tabular}
	}
	\vspace{-4mm}
\end{table}
\subsection{Diagonal Walking on a Flat Terrain } 
This scenario is focused on generating a diagonal walking reference trajectories for the simulated robot. In this scenario, the robot is standing in place (DS phase) and should take five steps according to the step input parameters that are shown in Table~\ref{tb:Params}. The simulation results are depicted in Fig.~\ref{fig:diag_walking}. In this simulation, the jerk commands are bounded between $\pm1.5 \frac{m}{s^3}$ and its maximum and minimum rate of change are considered to be $0.1$ and $-0.1$ respectively. The results showed that the proposed planner is able to generate stable walking with mostly stretched knees and keep the actual ZMP inside the support polygon. A video of this simulation is available online at \small{\url{https://www.dropbox.com/s/e5f3onmf4l0naj3/DiagonalWalking_HQ.mp4?dl=1}}.

\begin{figure}[!t]
	\centering
	\begin{tabular}	{c c}			
		\includegraphics[width=0.45\linewidth, trim= 0cm 0cm 0cm 0cm,clip]{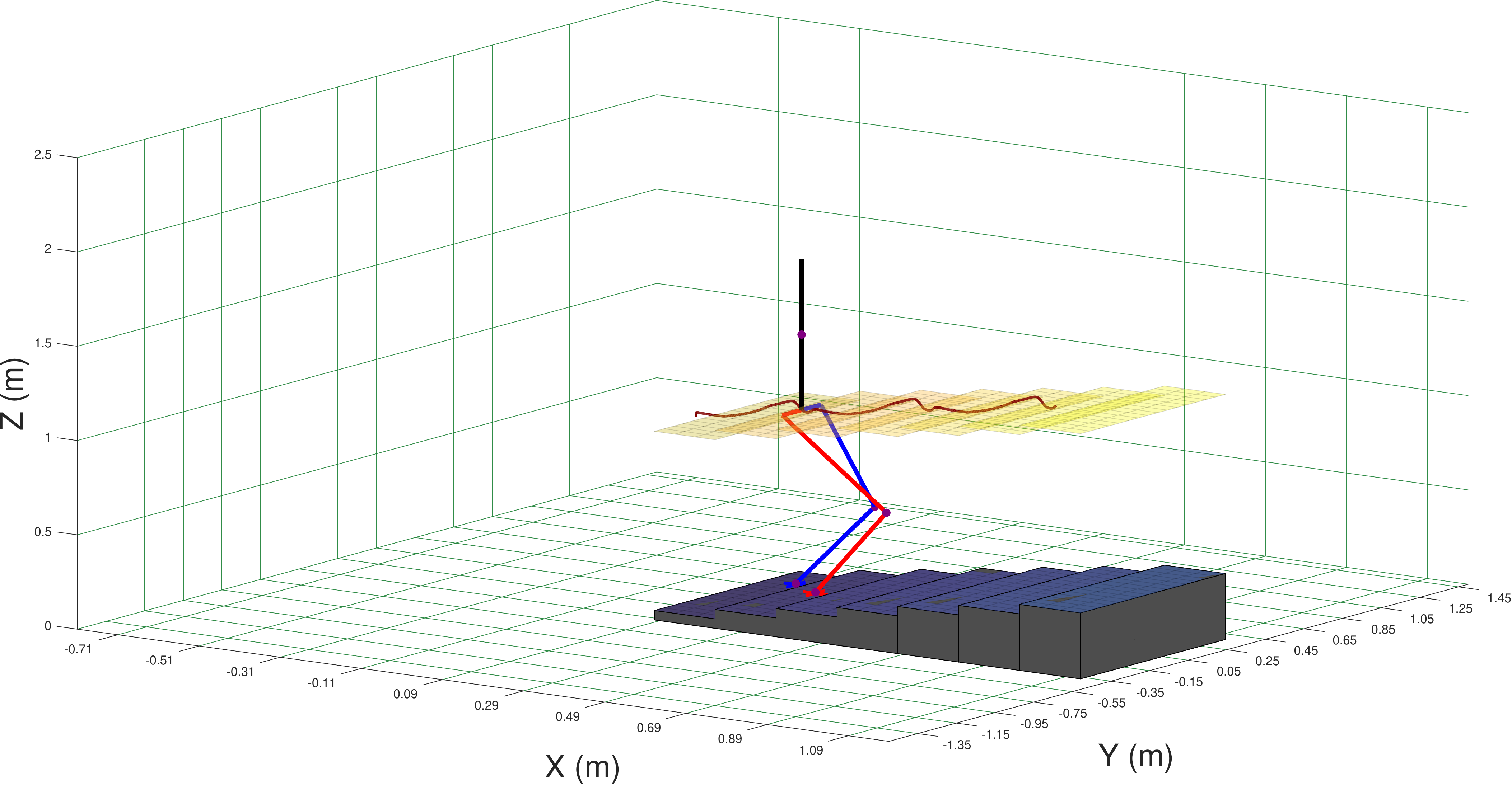}&
		\includegraphics[width=0.45\linewidth, trim= 0cm 0cm 0cm 0cm,clip]{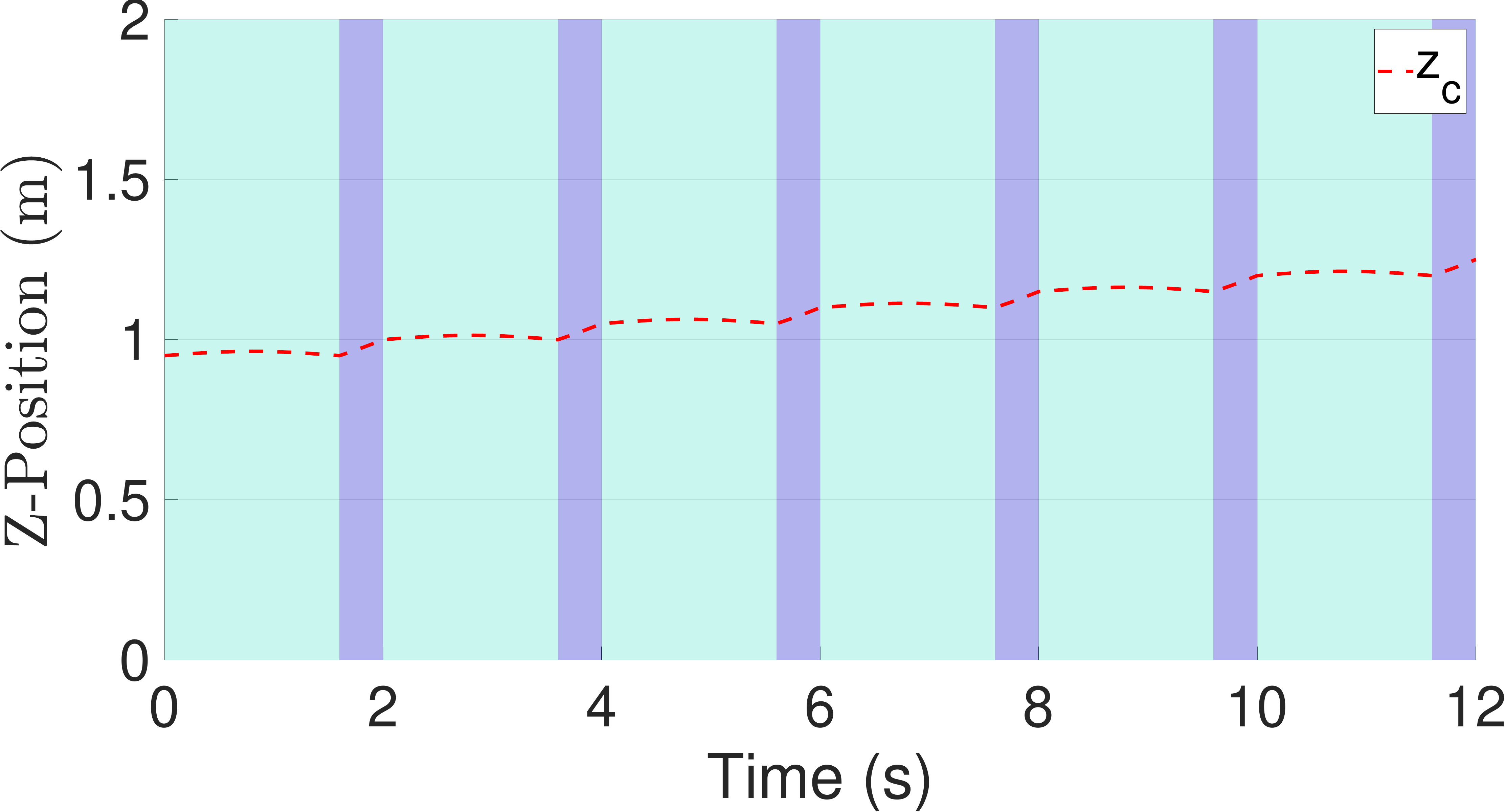}\\
		\includegraphics[width=0.45\linewidth, trim= 0cm 0cm 0cm 0cm,clip]{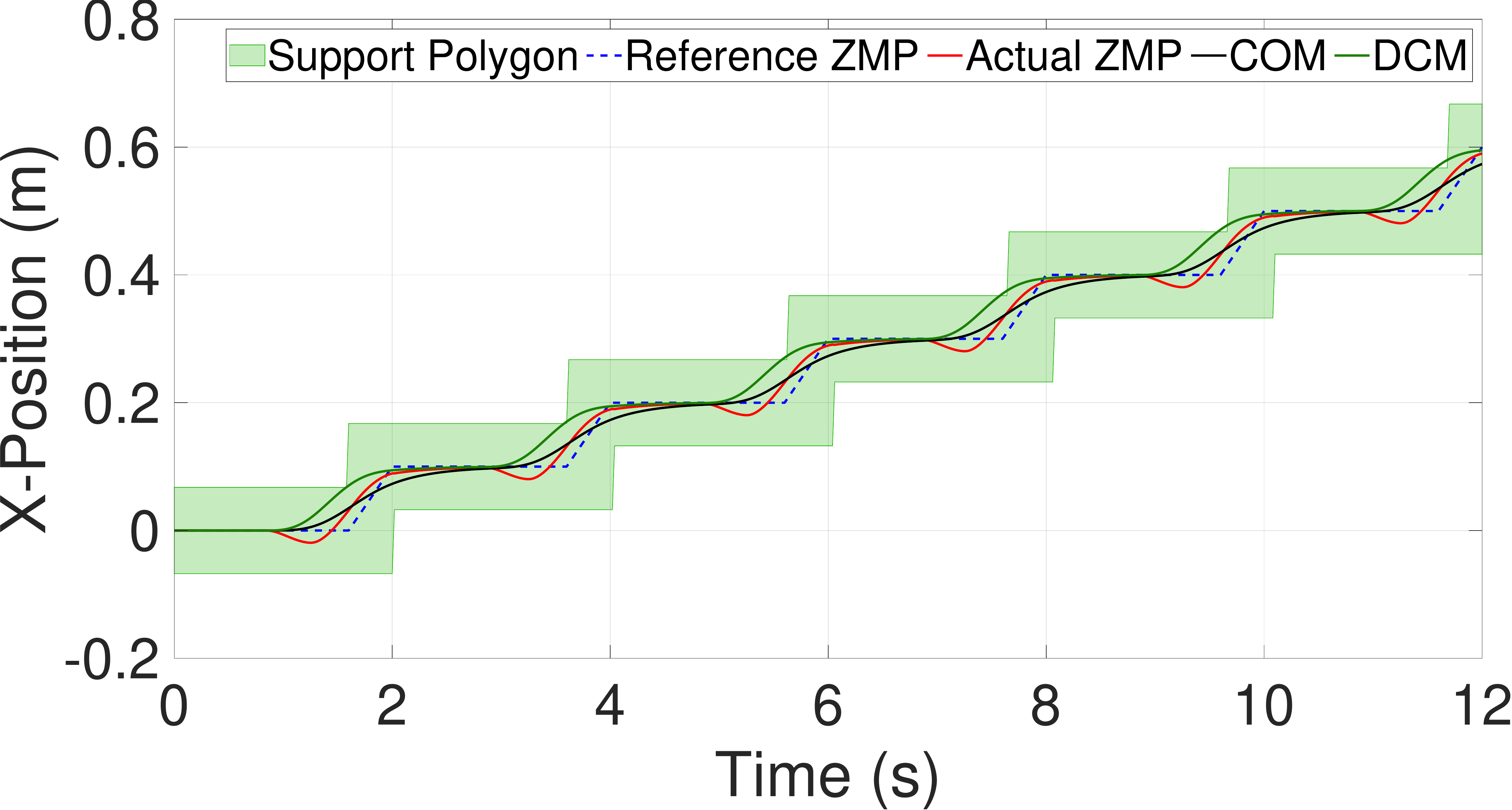}&
		\includegraphics[width=0.45\linewidth, trim= 0cm 0cm 0cm 0cm,clip]{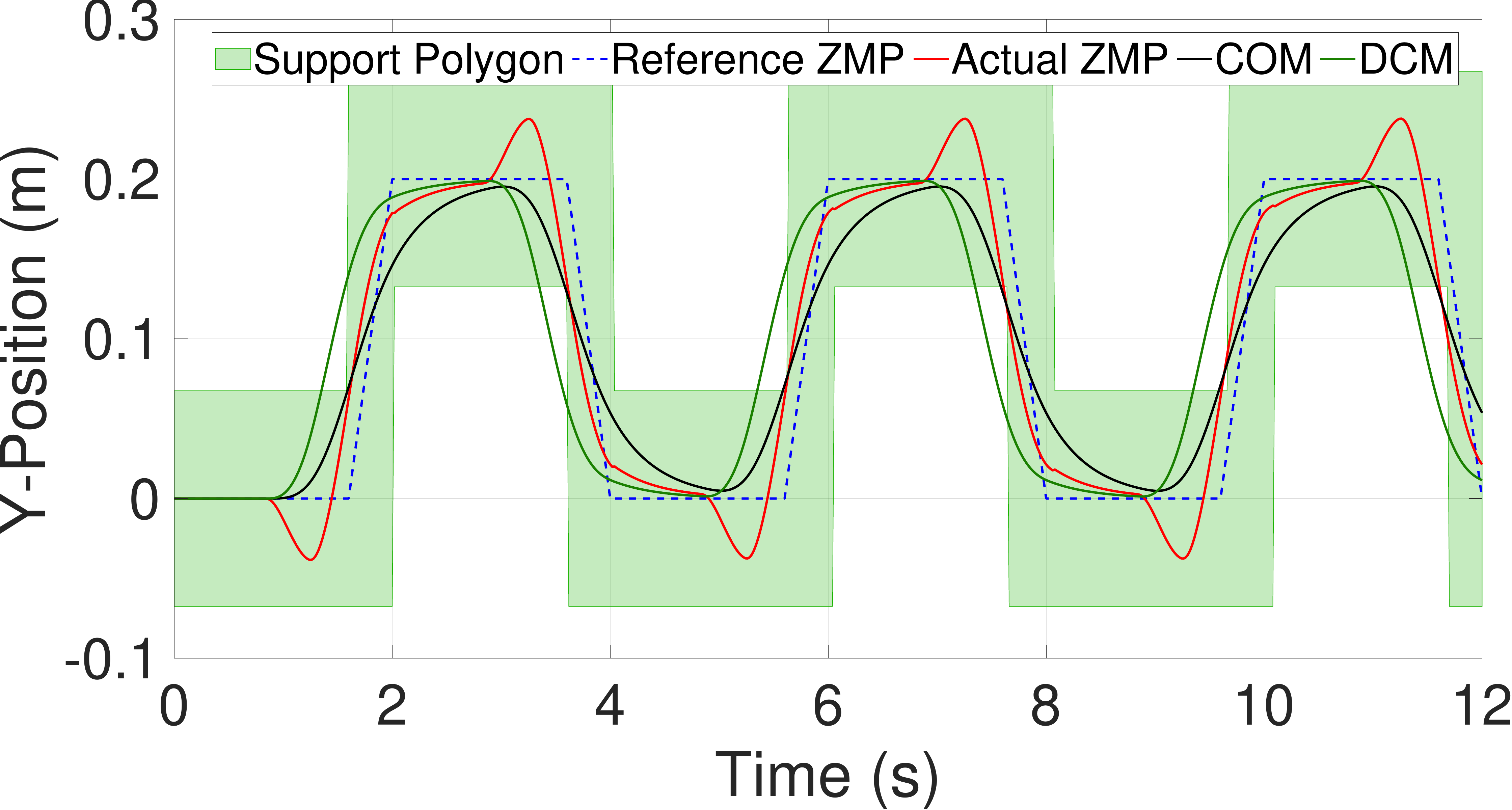}\\
	\end{tabular}	
	\vspace{-1.5mm}
	\caption{ The simulation results of stair climbing scenario: \textbf{top-left}: a snapshot of the stair climbing scenario; \textbf{top-right}: vertical reference trajectory of COM during stair climbing: cyan regions represent SS phase and violet regions represent DS phase; \textbf{bottom-left}: ZMP and COM trajectories in X-direction; \textbf{bottom-right}: ZMP and COM trajectories in Y-direction. }			
	\label{fig:stair_climbing}
		\vspace{-2mm}
\end{figure}

\subsection{Stair Climbing}
In our second scenario, to show the maneuverability of the proposed method, a stair climbing scenario is designed. In this scenario, the height of each stair is considered to be $0.1m$ and the simulated robot is assumed to be able to detect and determine the geometry of the stairs, therefore, it can plan a set of reliable places for foot placements. According to these assumptions, using (\ref{eq:zmpEquation}) and (\ref{eq:z_traj}), the trajectory of the ZMP and the vertical motion of the COM will be planned in advance. Afterward, the proposed TVMPC uses these trajectories as its references for planning the stair climbing locomotion. It should be noted that the proposed TVMPC considers the constraints that were explained in the previous scenario. The simulation results are depicted in Fig.~\ref{fig:stair_climbing}. The reference and the actual trajectories in $X$ and $Y$ direction are shown in the plots of the second row of this figure. As the results showed, the proposed method is capable to generate stable stair climbing locomotion. A video of this simulation is available online at \small{\url{https://www.dropbox.com/s/9xmdcxpkkjn3ujs/StairClimbing_HQ.mp4?dl=1}}.

\subsection{Robustness w.r.t. Measurement Noise}
\begin{figure}[!t]
	\centering
	\begin{tabular}	{c c }			
		\includegraphics[width=0.45\columnwidth, trim= 0cm 0cm 0cm 0cm,clip]{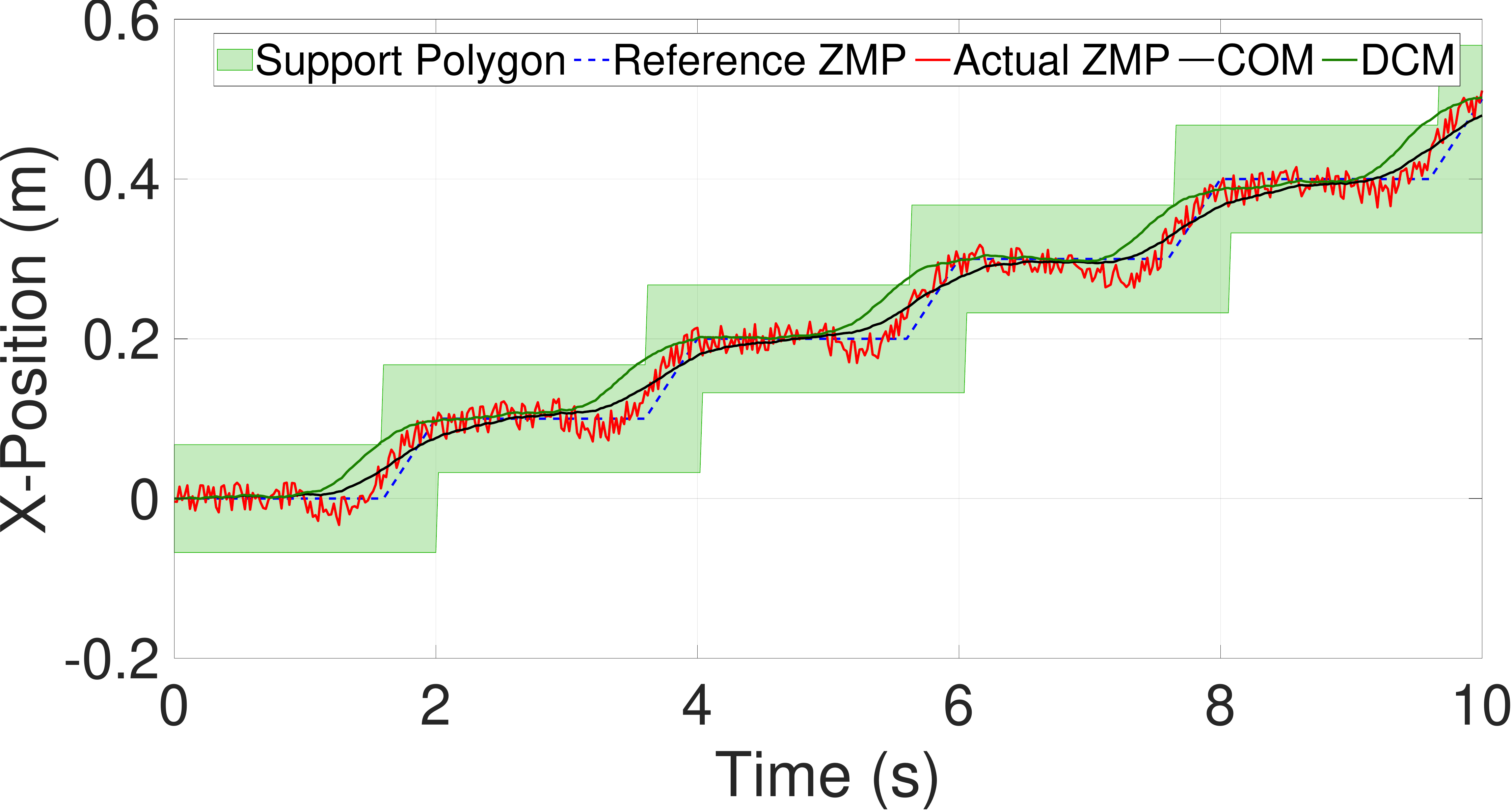}&
		\includegraphics[width=0.45\columnwidth, trim= 0cm 0cm 0cm 0cm,clip]{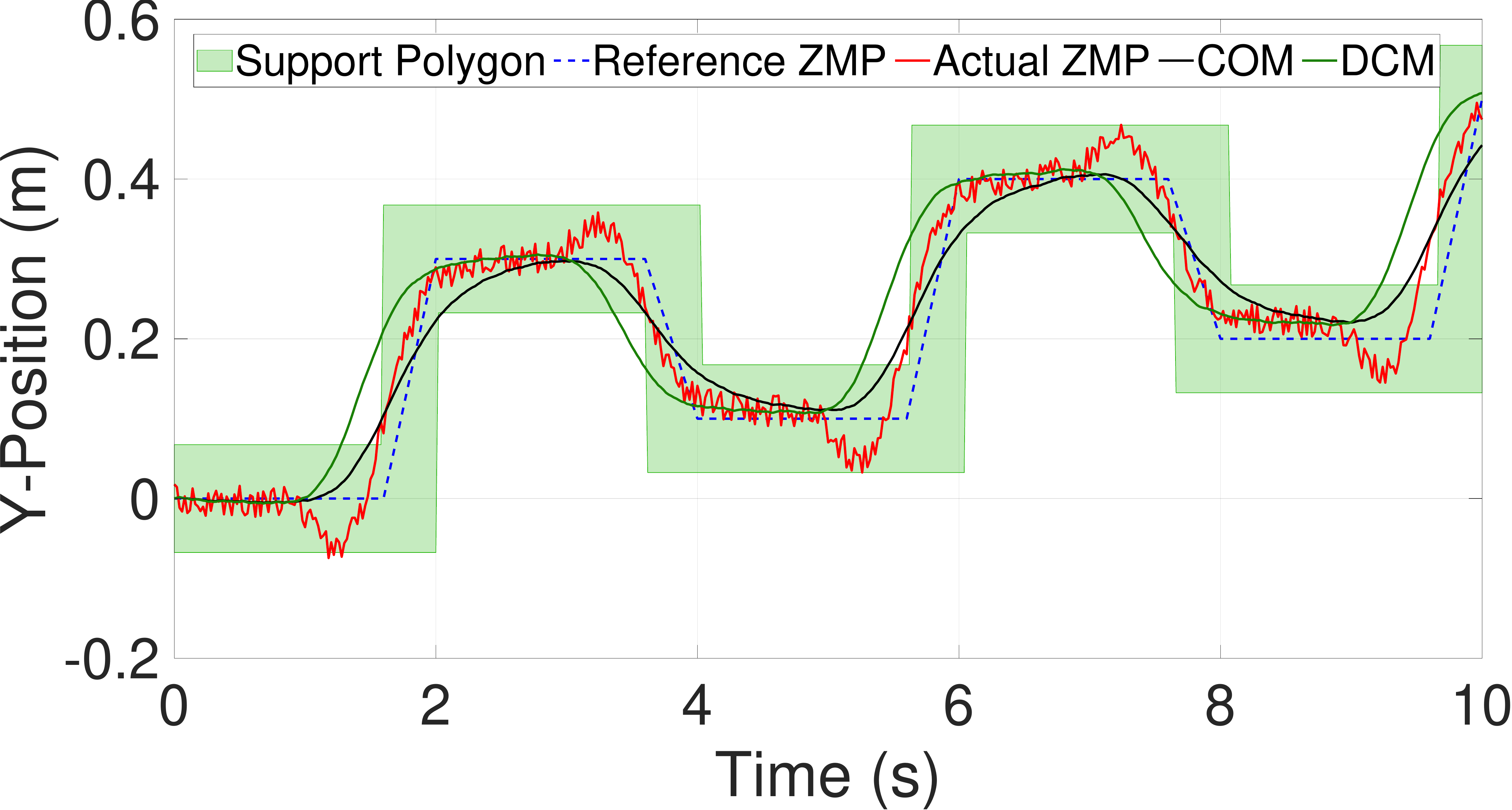}\\
		\includegraphics[width=0.45\columnwidth, trim= 0cm 0cm 0cm 0cm,clip]{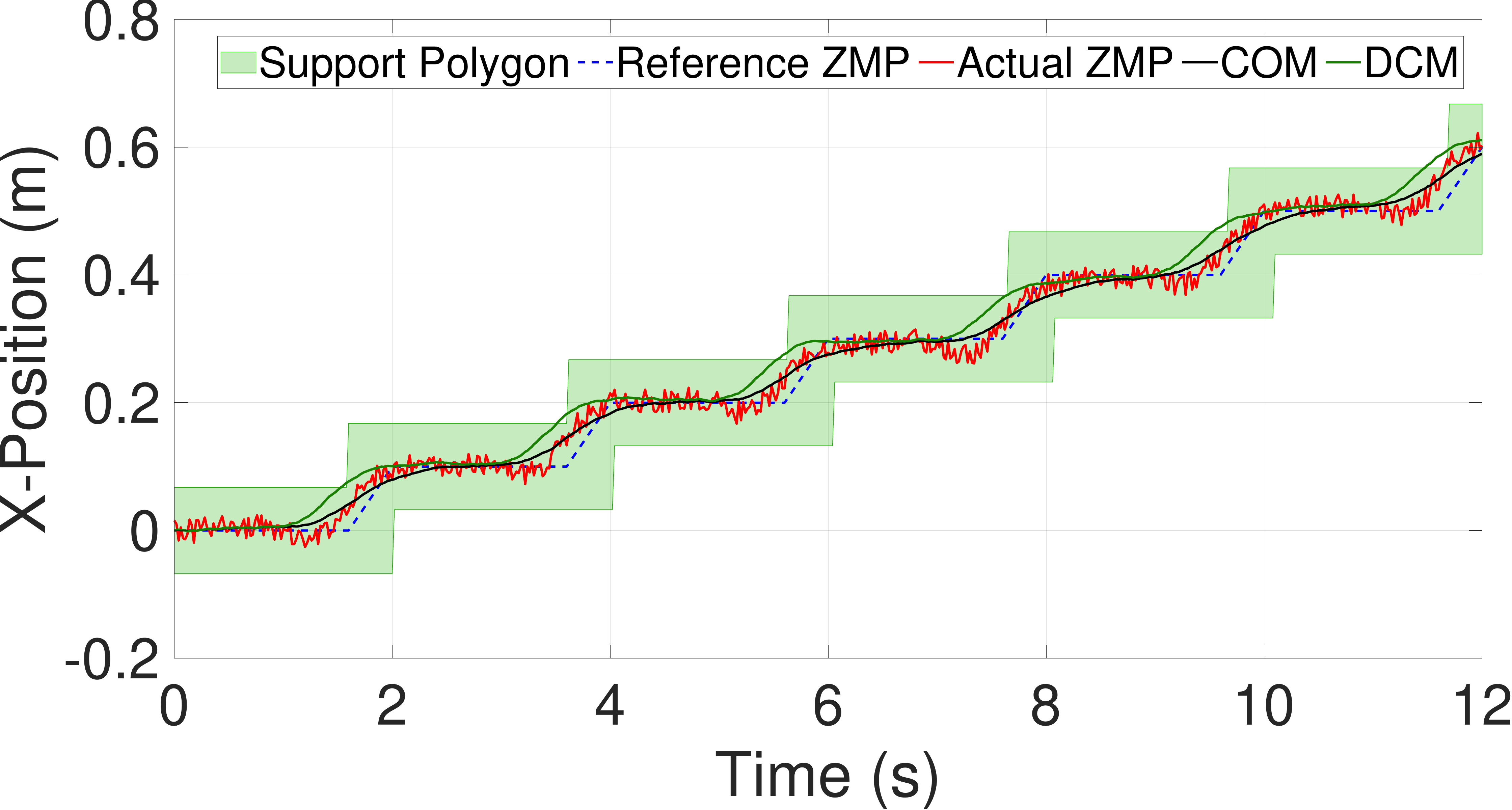}&
		\includegraphics[width=0.45\columnwidth, trim= 0cm 0cm 0cm 0cm,clip]{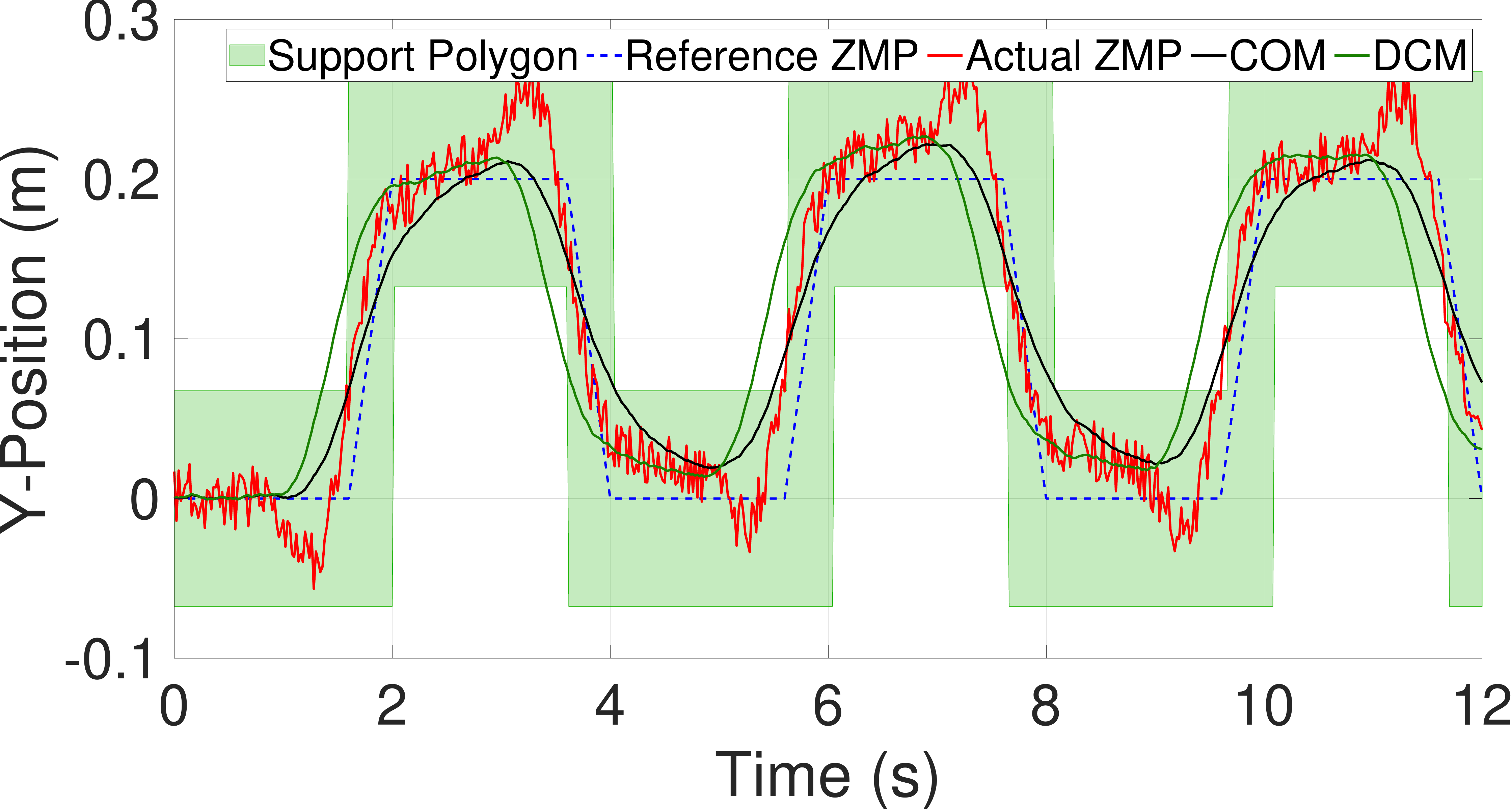}
	\end{tabular}	
	\vspace{-2mm}
	\caption{ The simulation results of examining the robustness w.r.t. measurement noise. The measurement is modeled as a stochastic process by adding a Gaussian noise ($-0.02m \le v \le 0.02m$) to the system output. The first row represents the results of the diagonal walking scenario and the second row represents the results of stair climbing scenario.}			
	\label{fig:robust_noise}
\end{figure}

In the real world, some reasons like noisy outputs of the sensors and simplifications in the dynamics model cause that the measurements are never perfect and generally affected by noise. A robust walking planner should be able to compensate this type of uncertainty and plan the reference trajectories robustly. As it is mentioned in Section~\ref{sec:DynamicsModel}, our planner uses a KF to estimate the states of the system. To examine the robustness of the proposed planner with regarding to measurement noise, the diagonal walking and stair climbing scenarios have been repeated in the presence of measurement noise. In these simulations, to simulate measurement noise, a zero-mean Gaussian noise~($v$) is added to the measurement. The simulations results are shown in Fig.~\ref{fig:robust_noise}. According to these results, our proposed planner is robust against this type of noise.

\subsection{Robustness w.r.t. COM Height Estimation Error}
\begin{figure}[!t]
	\centering
	\begin{tabular}	{c c }			
		\includegraphics[width=0.45\columnwidth, trim= 0cm 0cm 0cm 0cm,clip]{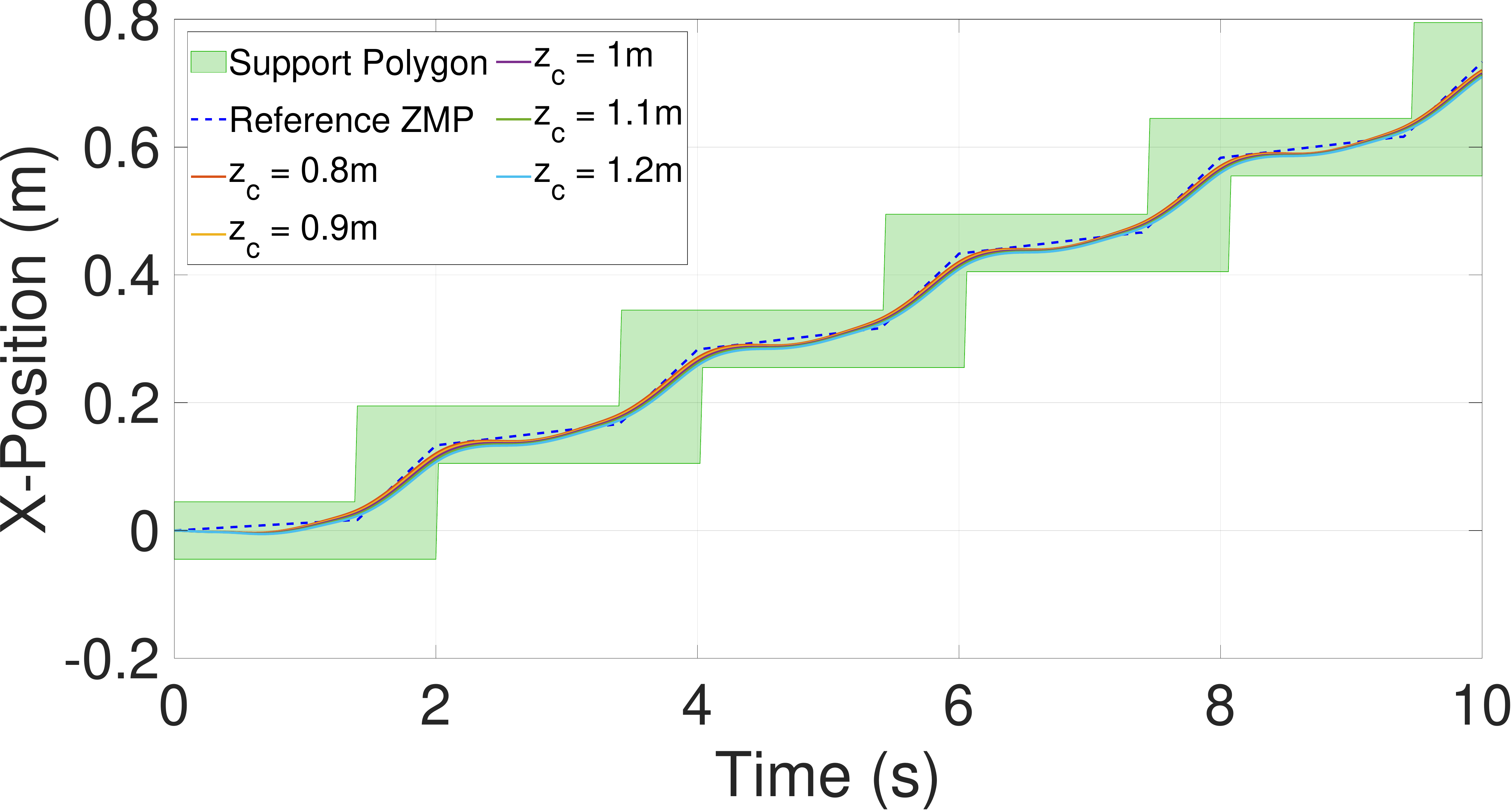}&
		\includegraphics[width=0.45\columnwidth, trim= 0cm 0cm 0cm 0cm,clip]{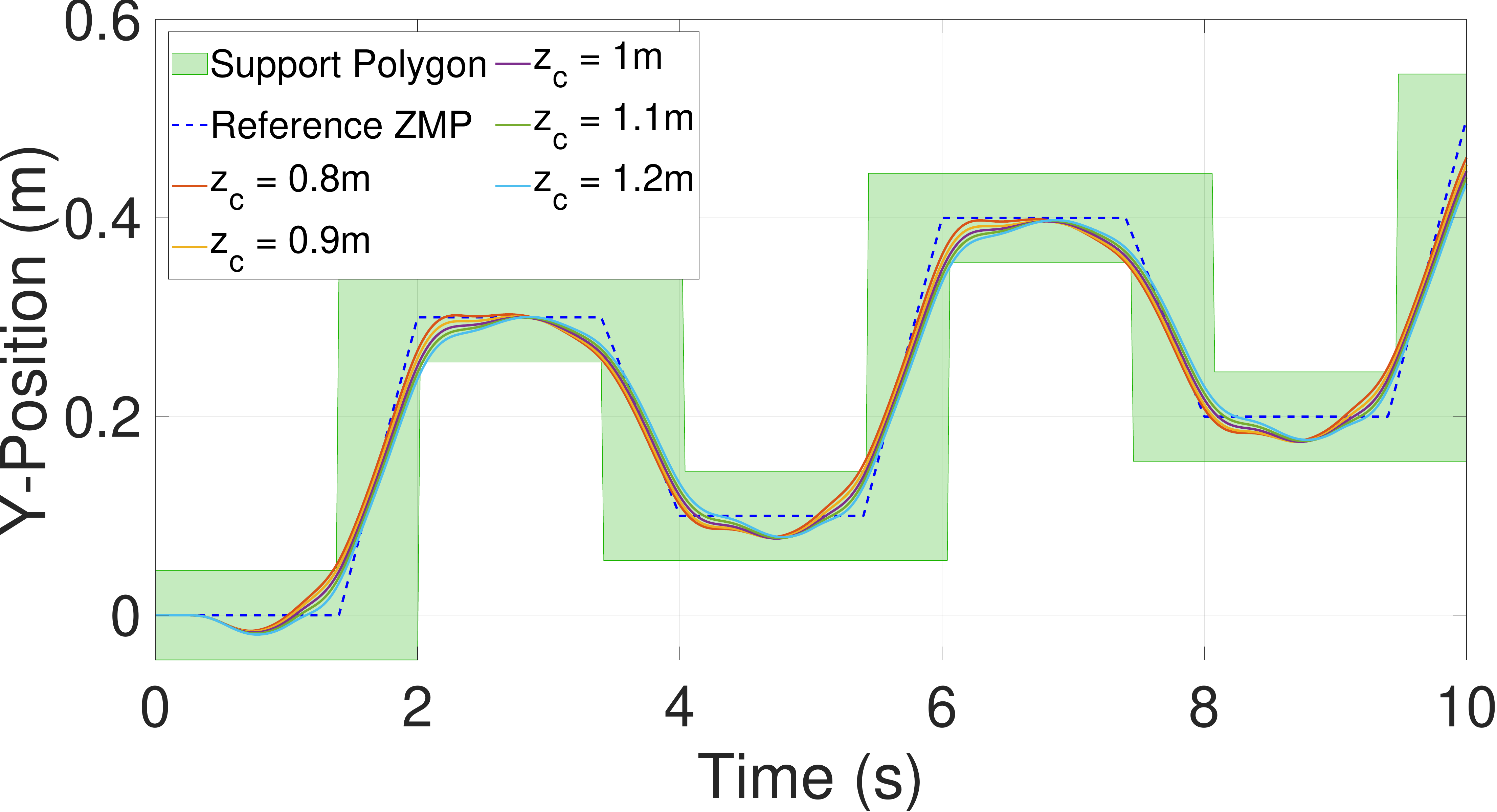}
	\end{tabular}	
	\vspace{-2mm}
	\caption{ The simulation results of examining the robustness w.r.t. COM height estimation error. Each curve represents the real ZMP (not the measured ZMP).}			
	\label{fig:robust_height}
	\vspace{-5mm}
\end{figure}

In some situations, like when a robot is carrying a heavy unknown weight object, the estimation of the height of COM is not accurate and the error may affect the performance of the planner. To examine the effect of this error, the diagonal walking scenario is repeated. In this simulation, the height of COM is considered to be in $z_c=1m$ for generating the reference trajectories and also for previewing the dynamics model but the actual height of the simulated robot is considered to be different. This simulation has been repeated four times with different errors~\mbox{($z_c = z_c\pm 0.1m , z_c = z_c\pm 0.2m$)}. The real ZMP in the sagittal and frontal planes have been recorded during the simulations and are shown in Fig.~\ref{fig:robust_height}. Based on the recorded ZMPs, unlike the sagittal plane, COM height estimation error has a significant effect in the frontal plane. Moreover, the simulation result showed that the controller could track the inaccurate reference trajectories and generate stable walking by keeping the ZMP inside the support polygon.

\begin{figure*}[!t]
	\centering
	\begin{tabular}	{c c c c}			
		\includegraphics[width=0.45\columnwidth, trim= 0cm 0cm 0cm 0cm,clip]{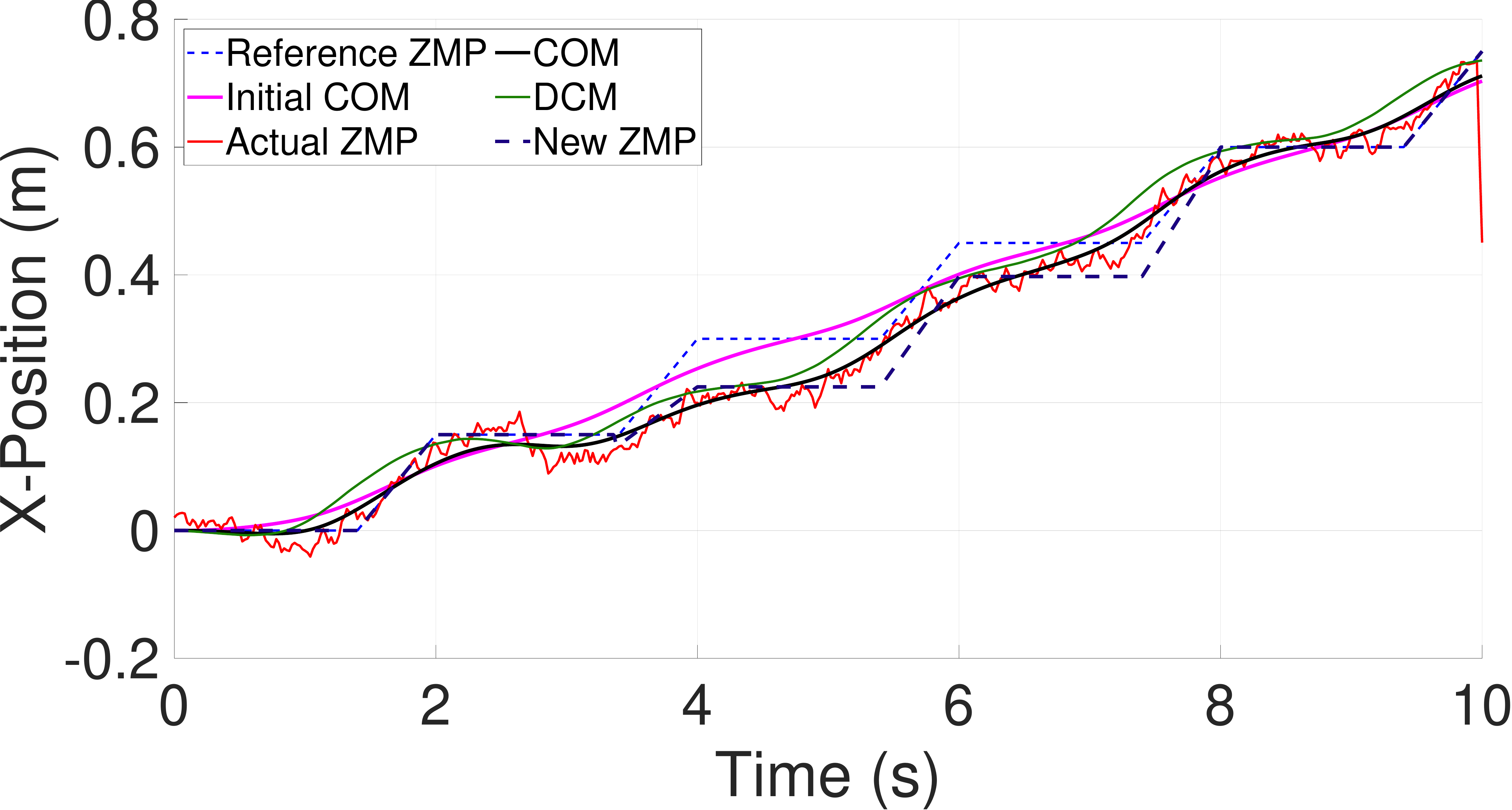}&
		\includegraphics[width=0.45\columnwidth, trim= 0cm 0cm 0cm 0cm,clip]{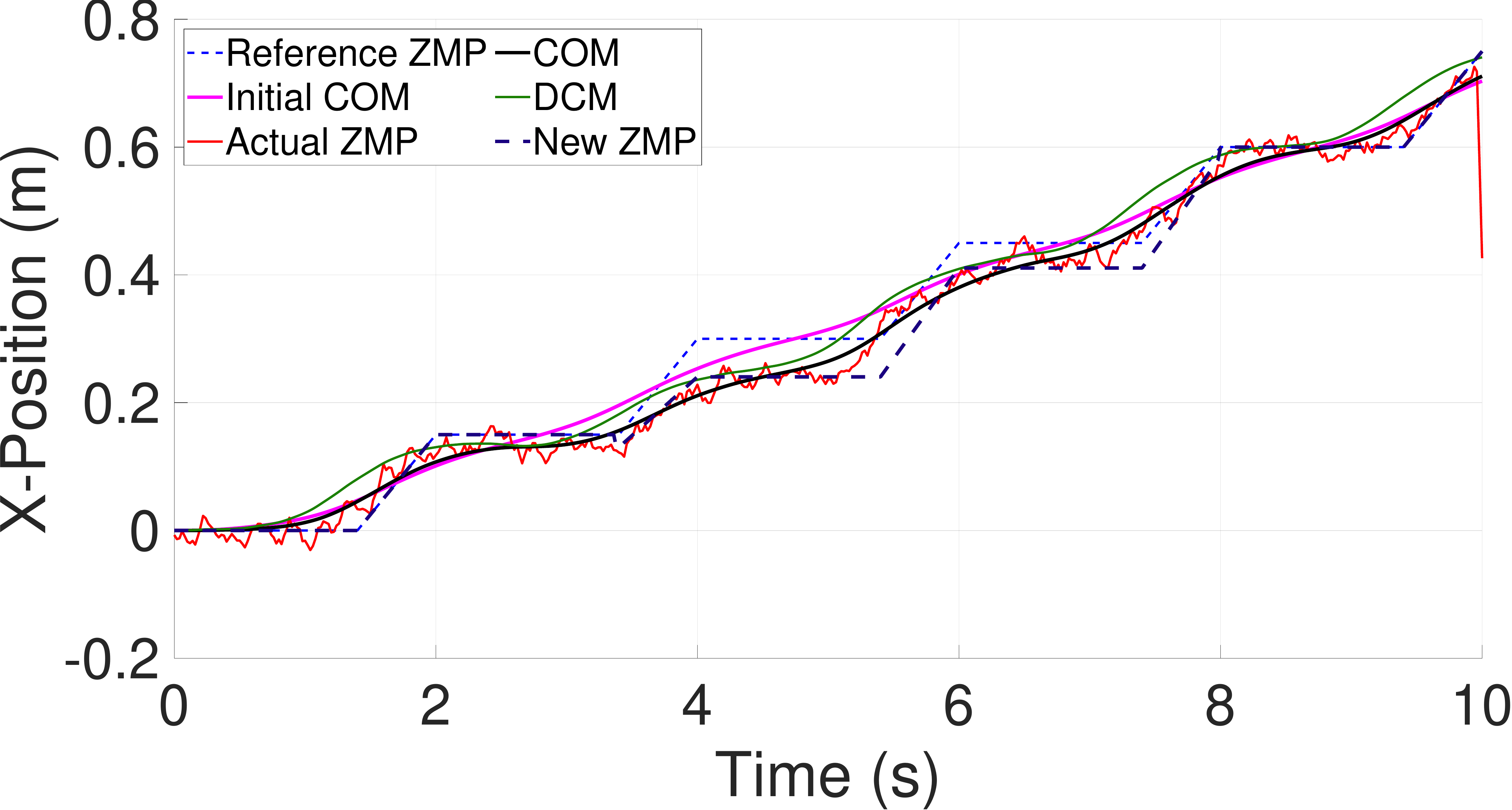}&
		\includegraphics[width=0.45\columnwidth, trim= 0cm 0cm 0cm 0cm,clip]{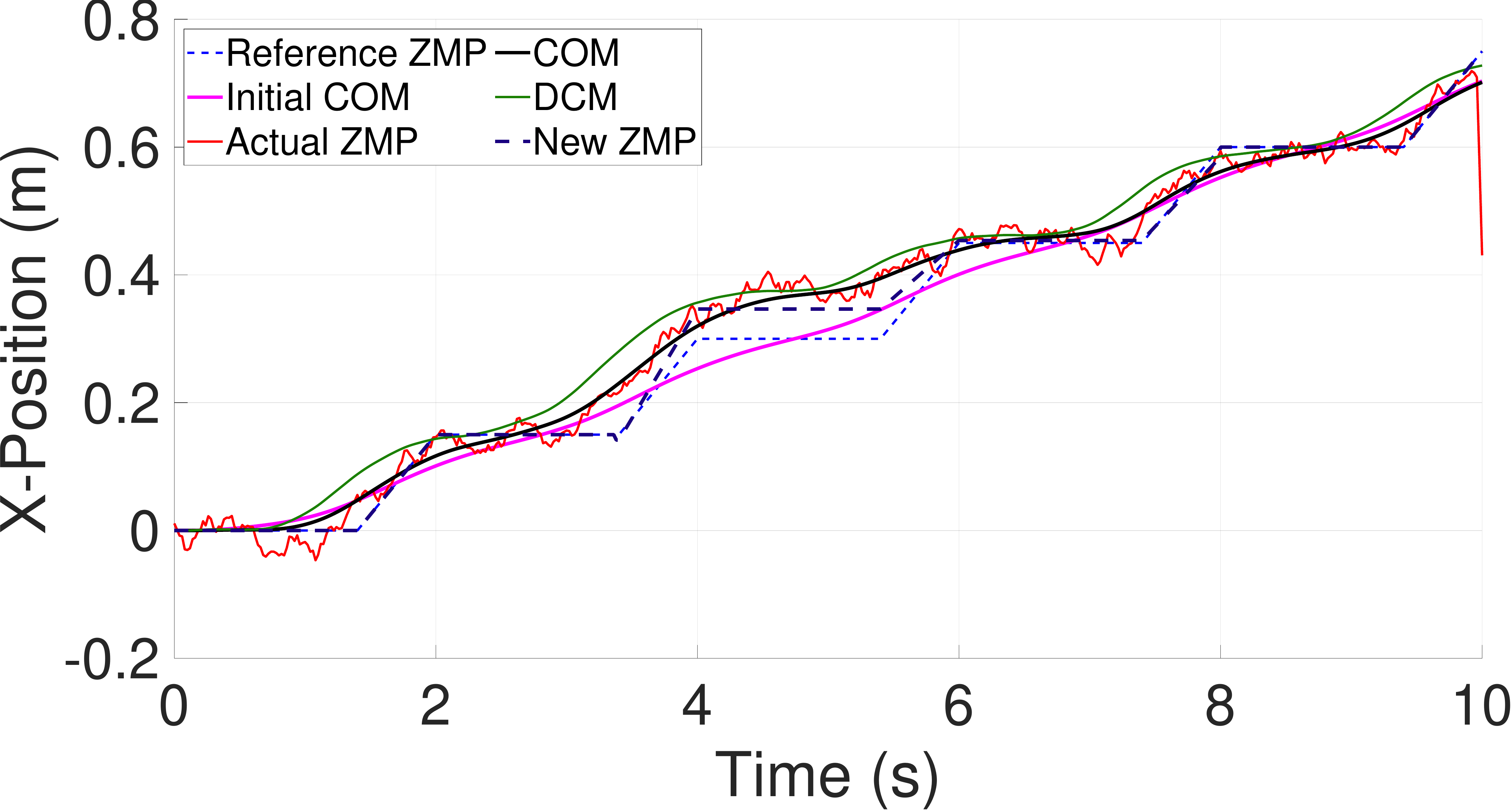}&
		\includegraphics[width=0.45\columnwidth, trim= 0cm 0cm 0cm 0cm,clip]{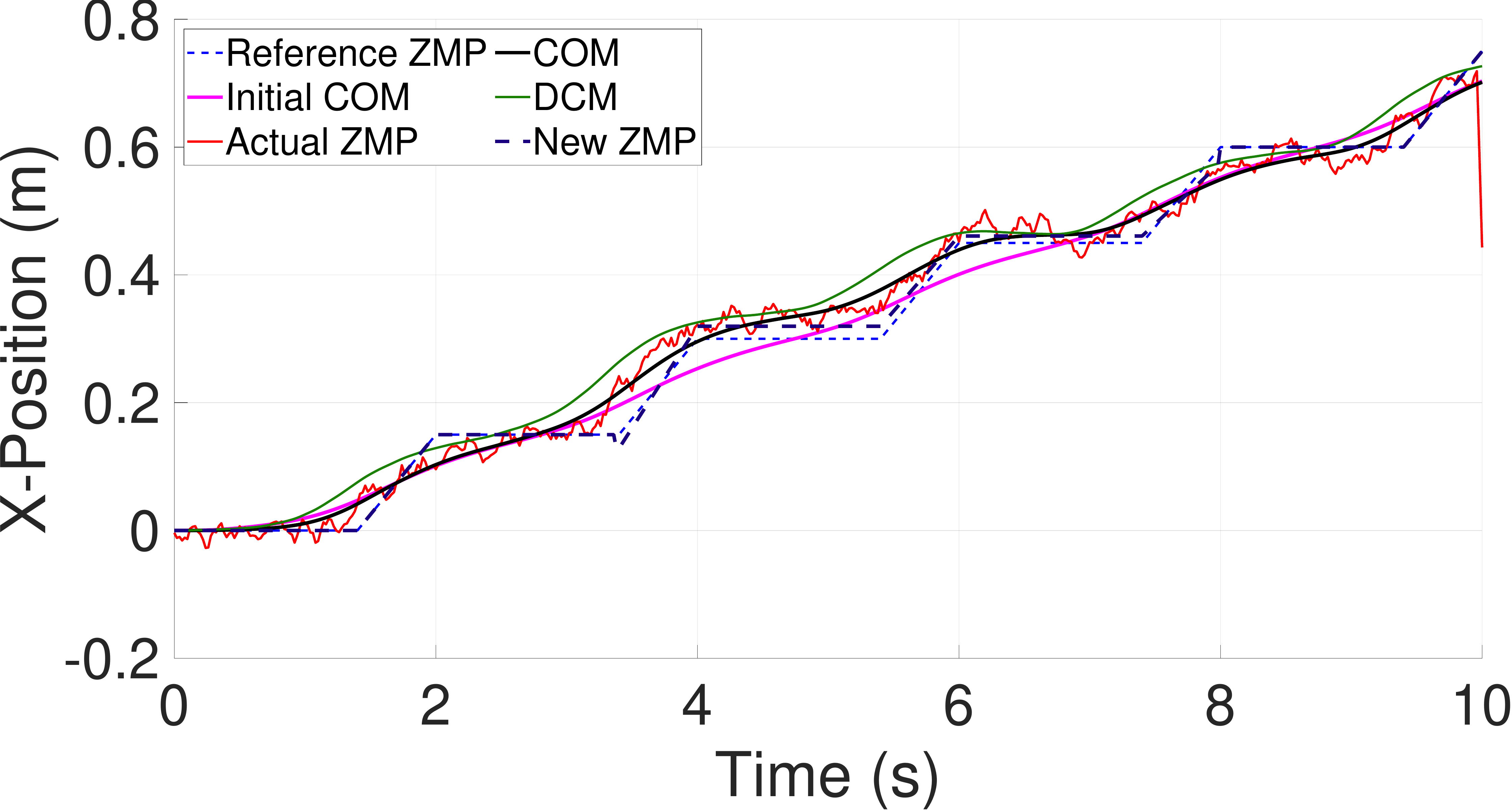}\\
		
		\includegraphics[width=0.45\columnwidth, trim= 0cm 0cm 0cm 0cm,clip]{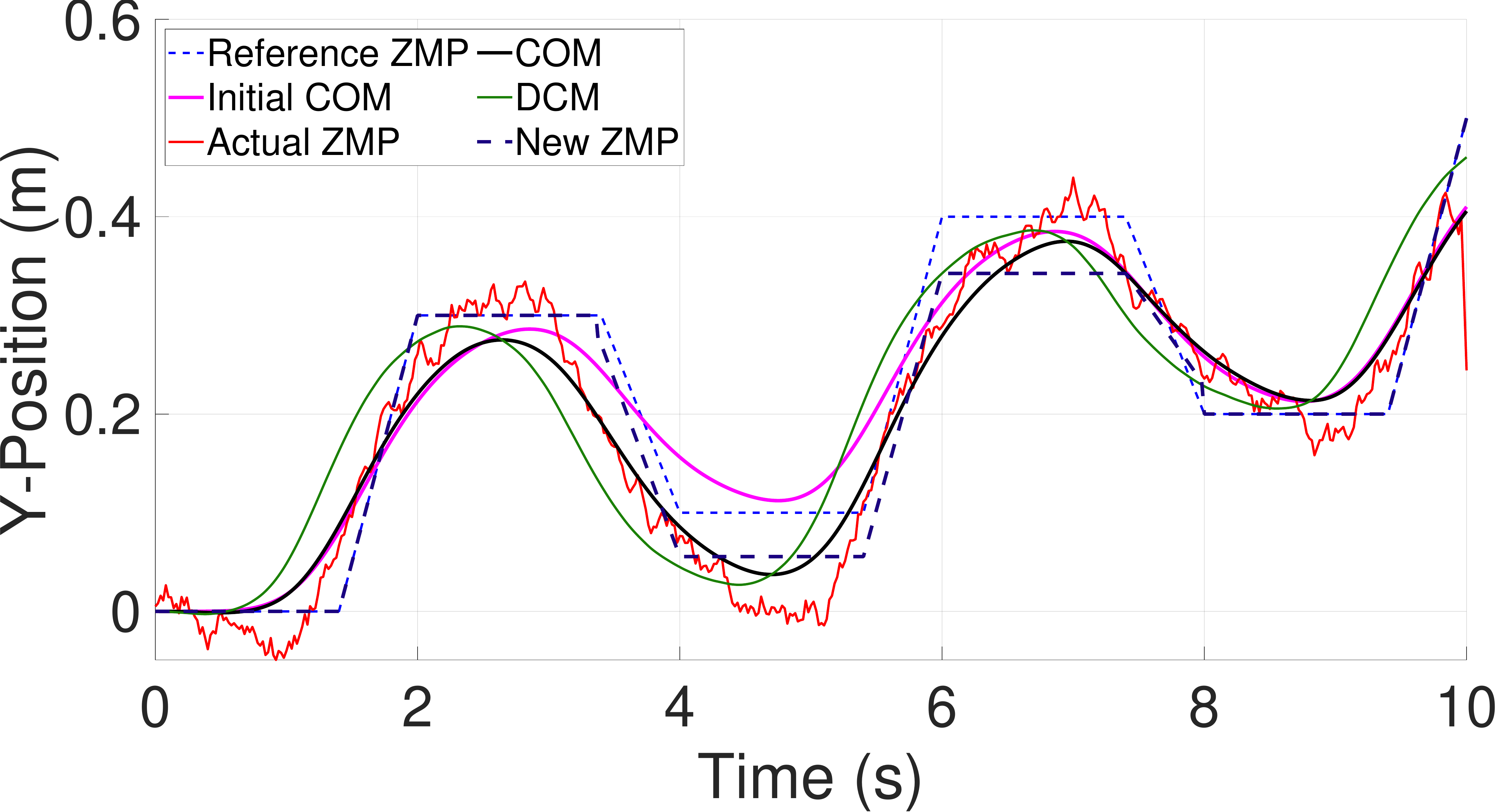}&
		\includegraphics[width=0.45\columnwidth, trim= 0cm 0cm 0cm 0cm,clip]{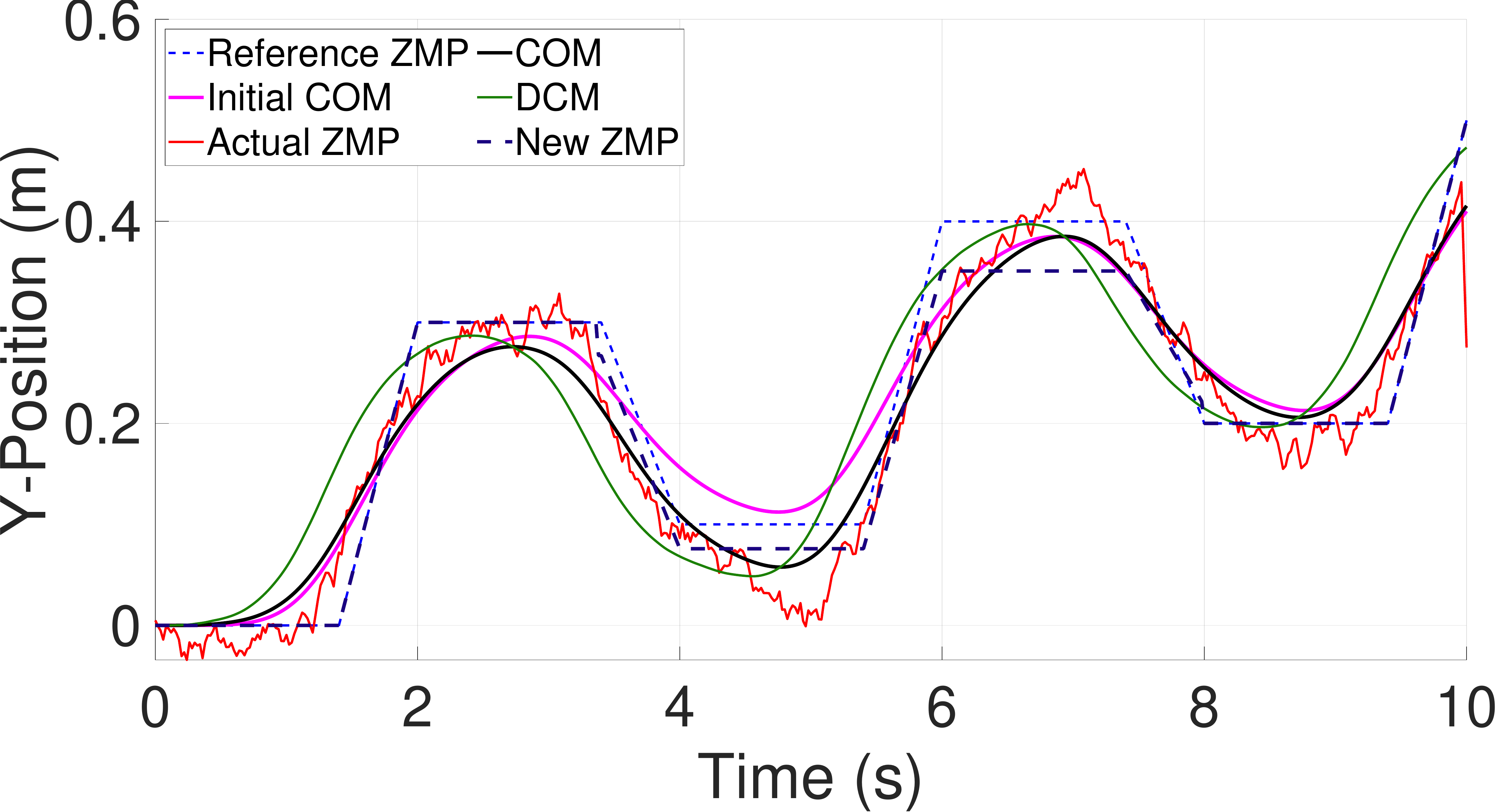}&
		\includegraphics[width=0.45\columnwidth, trim= 0cm 0cm 0cm 0cm,clip]{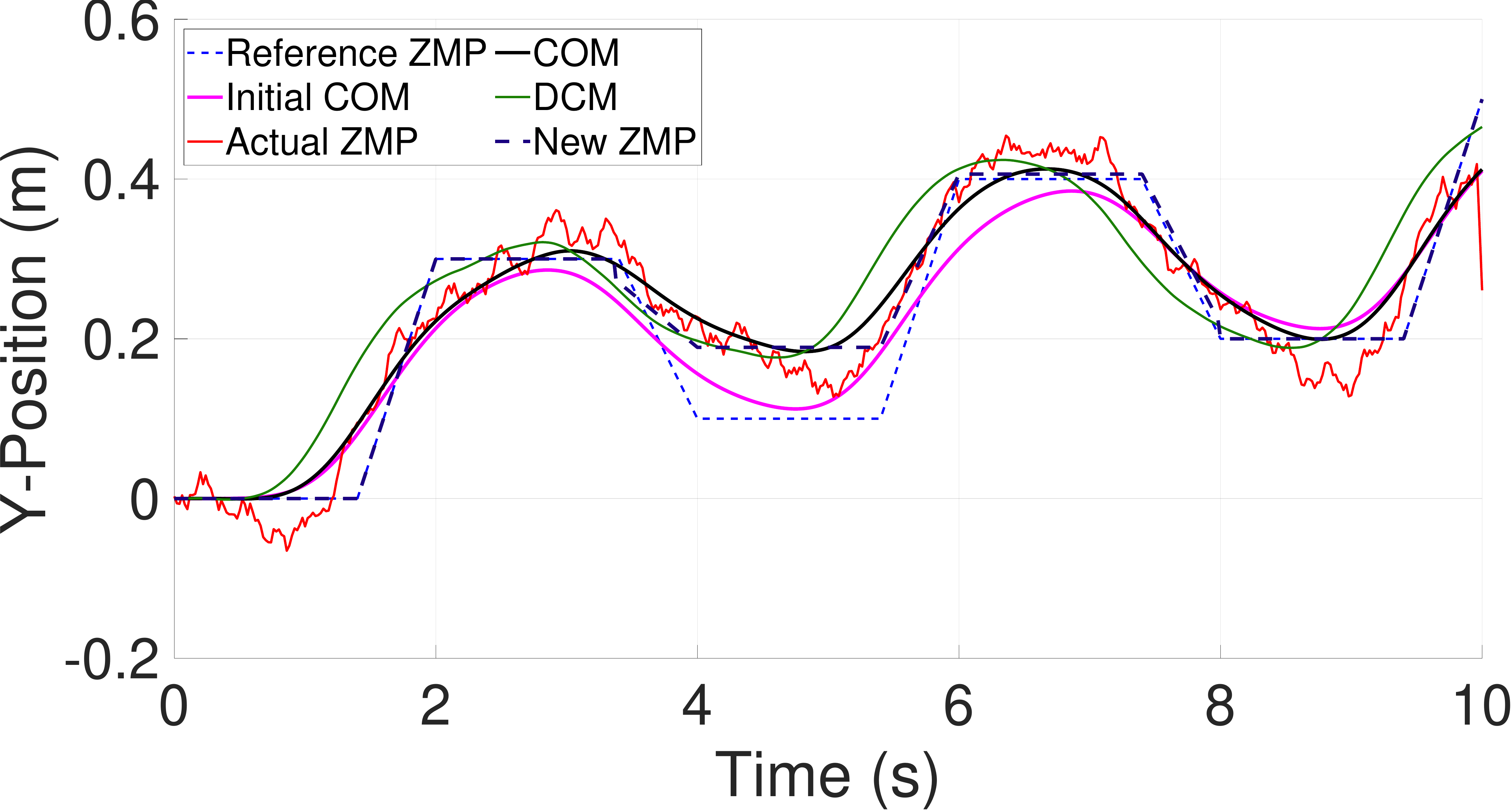}&
		\includegraphics[width=0.45\columnwidth, trim= 0cm 0cm 0cm 0cm,clip]{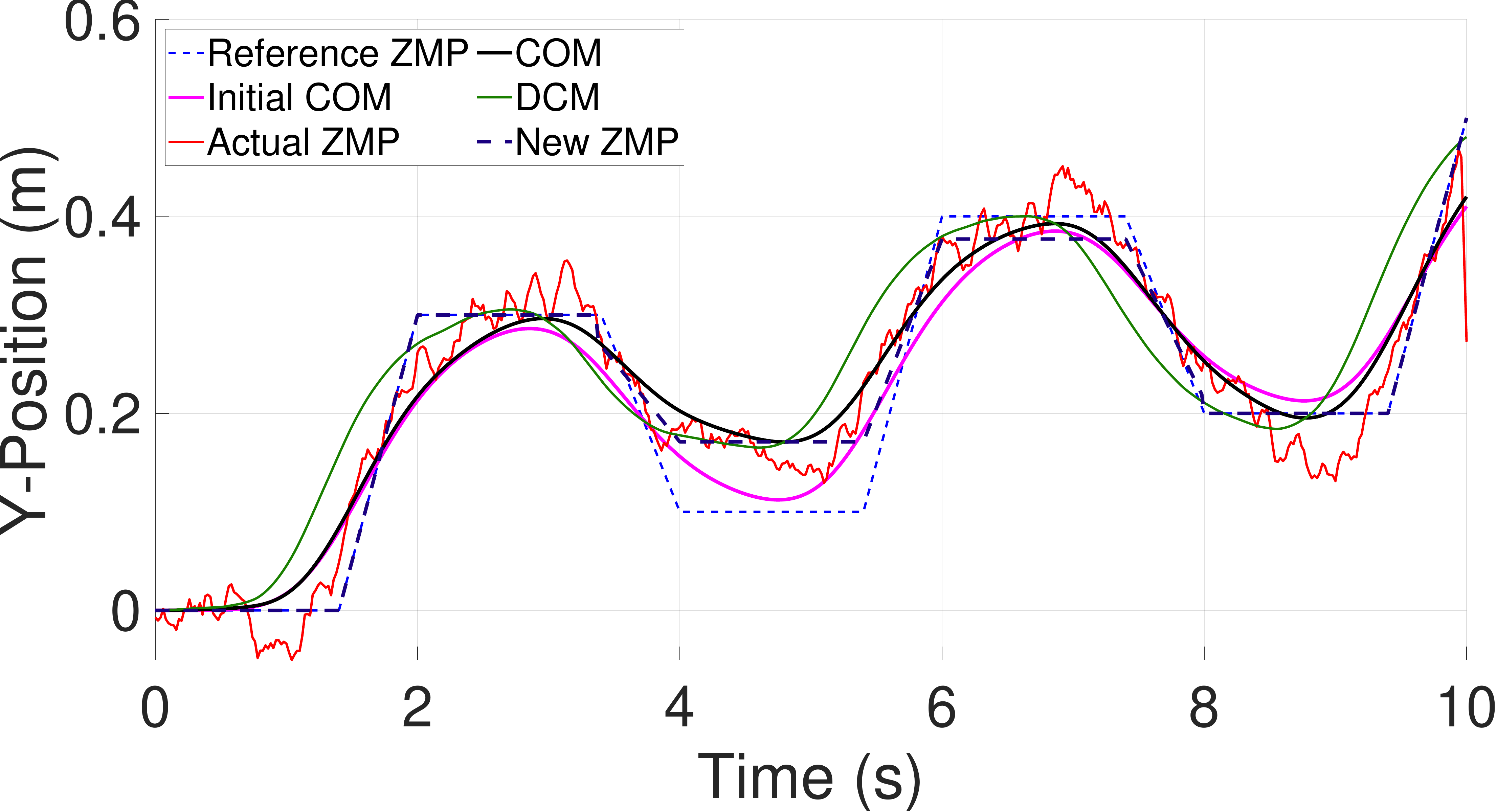}\\
		
		\small{$F_x = F_y = -75 N $}&\small{$F_x = F_y  = -50 N$}&\small{$F_x = F_y = 75 N $}&\small{$F_x = F_y = 50 N $}
	\end{tabular}	
	\caption{ The simulation results of examining the robustness w.r.t. external disturbances. After applying a disturbance, the proposed planner modifies the ZMP and the COM reference trajectory to regain the stability of the robot.}		
	\vspace{-7mm}	
	\label{fig:robust_ext}
\end{figure*}

\subsection{Robustness w.r.t. External Disturbance}
In some conditions like a collision with an obstacle or a push by someone, the low-level controller should be able to react against these disturbances and guarantee the tracking performance to keep the stability of the robot. Generally, the low-level controller tries to cancel the effect of these disturbances by applying compensating torques at the ankle, hip and arms joints. In the case of severe disturbances, the low-level controller could not negate the effect of disturbance because of the saturation of the joint torques. One of the most important abilities of an online robust walking planner is adjusting the reference plan in order to return ZMP inside the support polygon and keep the stability of the robot.
To analyze the robustness of the proposed planner w.r.t. unpredictable external disturbance, a set of simulations has been carried out. In these simulations, the simulated robot should walk diagonally and, while it is walking, an external force will be applied at its COM in $t = 2.2s$ with an impact duration of $\Delta t = 10ms$. 
It should be noted that in order to be more realistic, these simulations have been performed in the presence of measurement noise~($-0.02m\le v \le0.02m$). The simulation has been repeated four times with different push amplitude and direction. Indeed, in each trial, the impact duration has been fixed and the amplitude and the direction of the impact have been changed~(from $-75N$ to $+75N$). The simulation results are depicted in Fig.~\ref{fig:robust_ext}. In this figure, each column represents the result of a single simulation. As are shown in these plots, once a disturbance is applied, the reference planner modifies the planned COM trajectory online to manage the effect of disturbance. According to the simulation results, the proposed planner is robust against external disturbance and can modify online the plan to handle such situations.
\section{Discussion and Comparison}
\label{sec:discussion}
According to the simulation results presented in the previous section, the proposed planner is not only able to plan the locomotion trajectories optimally, but also it is robust against uncertainties and disturbances. Unlike~\cite{kajita2003biped},~\cite{morisawa2005pattern},~\cite{shimmyo2013biped} and~\cite{kajita2018biped}, which are based on ZMP preview control, the proposed planner is able to modify online the landing location of the swing leg regarding DCM measurement. Additionally, unlike most of the works presented in the literature, the proposed planner takes into account the vertical motion of the COM and based on it can predict the future behavior of the dynamics model of the system. This prediction improves the performance of the locomotion planning by updating the model and the nominal conditions at each control interval. Besides, in our proposed architecture the low-level controller is decoupled from the planing layer to fade the complexity of the low level control, but the sensory information is adapted to perform online modification of the generated plan. 
\vspace{-2mm}
\section {Conclusion}
\label{sec:CONCLUSION}
In this paper we have tackled the problem of developing a robust closed-loop biped locomotion planner to generate walking with mostly stretched knee which is more human-like and efficient. The fundamental idea behind the proposed planner was formulating the planning problem as a TVMPC to be able to predict the future behavior of the dynamics model according to the vertical motion of the COM. Moreover, the proposed planner was able to find an optimal solution by considering some mixed constraints in the dynamics system states, inputs and outputs. Besides, the concept of DCM  was used to modify the landing location of the swing leg in an online manner to increase the robustness level. The performance of the proposed planner has been validated using two simulation scenarios which were \textit{diagonal walking on a flat terrain} and \textit{stair climbing}. In addition to these simulations, three simulations have been performed to analyze the robustness of the proposed planner regarding measurement noise, COM height estimation error and external disturbances. The simulation results validated the performance and robustness of the proposed planner.  In future work, we would like to investigate the effect of online modification of step duration and propose a method which considers it as a decision variable. 
\vspace{-2mm}
\section*{Acknowledgment}
This research is supported by Portuguese National Funds through Foundation for Science and Technology (FCT) through FCT scholarship SFRH/BD/118438/2016.

\bibliographystyle{IEEEtran}
\bibliography{ICRA2019}

\end{document}